\documentclass[10pt,twocolumn,letterpaper]{article}

\usepackage[pagenumbers]{cvpr} %

\usepackage{cuted} %

\usepackage{currfile} %

\usepackage{caption} %

\newcommand{\TODO}[1]{\textbf{\color{red}[TODO: #1]}}
\renewcommand{\TODO}[1]{}

\usepackage{amsthm}
\newtheorem{proposition}{Proposition}
\usepackage{multirow}
\usepackage{colortbl}
\usepackage{tikz}
\usetikzlibrary{arrows.meta,positioning,fit,backgrounds,calc,shapes.geometric}
\usepackage{pgfplots}
\pgfplotsset{compat=1.17}

\newcommand{\frameslot}[4]{\includegraphics[width=#2,height=#3]{figs/frames/#1}}

\newcommand{\ours}{PEA\xspace}          %
\newcommand{\oursfull}{PEA\xspace}      %
\newcommand{\pid}{PEA-ID\xspace}        %
\newcommand{\gid}{Gain-ID\xspace}       %
\newcommand{\raw}{Original\xspace}      %
\newcommand{\learn}{Learned-only\xspace} %
\newcommand{\teach}{Teacher\xspace}     %
\newcommand{\sg}[2]{#1\,/\,#2}

\AtBeginDocument{%
  \crefname{figure}{Figure}{Figures}\Crefname{figure}{Figure}{Figures}%
  \crefname{table}{Table}{Tables}\Crefname{table}{Table}{Tables}%
}

\usepackage{placeins} %

\definecolor{cvprblue}{rgb}{0.21,0.49,0.74}
\usepackage[pagebackref,breaklinks,colorlinks,allcolors=cvprblue]{hyperref}

\title{Probing Invisible Actuation Shifts for Test-Time Calibration of \\ Visuomotor Policies}

\author{
Jie Liu$^{1,*,\dagger}$ \qquad Hui Xie$^{1,*}$ \qquad Qinfeng Feng$^{2}$ \qquad Joaquin Vanschoren$^{1}$ \\[0.6em]
{\normalsize $^{1}$Eindhoven University of Technology \qquad $^{2}$The University of Hong Kong}
}

\begin{document}
\addtocontents{toc}{\protect\setcounter{tocdepth}{-1}}
\twocolumn[{\renewcommand\twocolumn[1][]{#1}%
\maketitle
\centering
\resizebox{\textwidth}{!}{%
\begin{tikzpicture}[font=\scriptsize, lab/.style={font=\scriptsize\bfseries, anchor=south west, inner sep=0pt}]
\node[anchor=south west, inner sep=0] (a1) at (0,0) {\frameslot{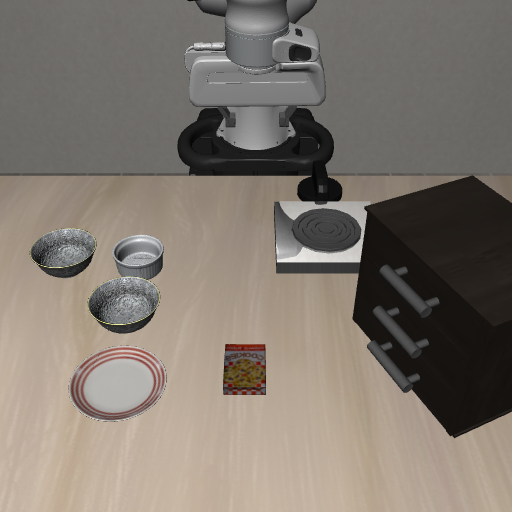}{2.9cm}{2.9cm}{agentview frame\\nominal camera\\(task start)}};
\node[anchor=south west, inner sep=0] (a2) at (3.05,0) {\frameslot{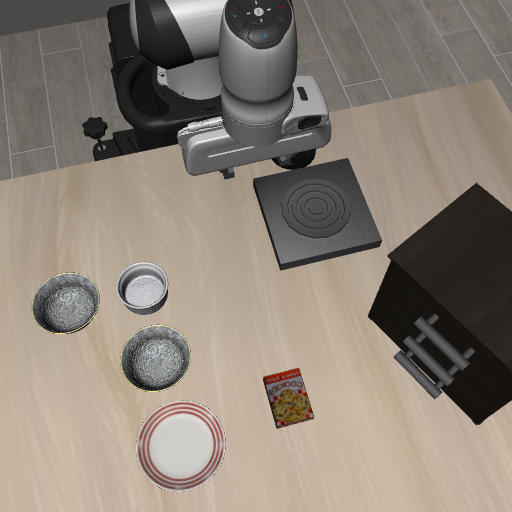}{2.9cm}{2.9cm}{same state,\\rotated camera\\(azimuth, elevation)}};
\node[font=\tiny, below=1pt of a1] {nominal};
\node[font=\tiny, below=1pt of a2] {camera moved};
\node[lab] at (0,3.05) {(a) Visual shift: the image changes};
\node[anchor=south west, inner sep=0] (b1) at (6.55,0) {\frameslot{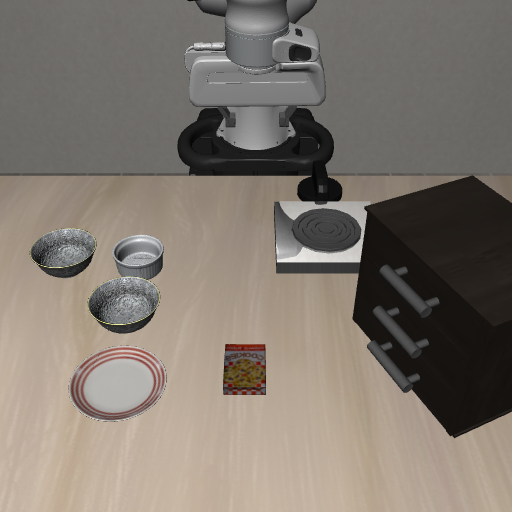}{2.9cm}{2.9cm}{agentview frame,\\nominal camera:\\identical under all\\four responses $\rightarrow$}};
\node[font=\tiny, below=1pt of b1] {what the policy sees};
\node[anchor=south west, inner sep=0] (b2) at (9.55,-0.45) {\includegraphics[height=3.35cm]{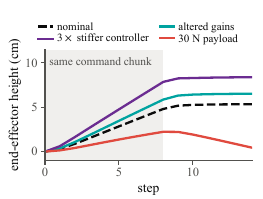}};
\node[lab] at (6.55,3.05) {(b) Actuation shift: same image, same command, different motion};
\draw[black!20] (14.0,-0.4) -- (14.0,3.2);
\node[anchor=south west, inner sep=0] (c) at (14.15,-0.45) {\includegraphics[height=3.35cm]{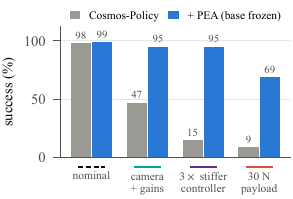}};
\node[lab] at (14.2,3.05) {(c) LIBERO-Spatial, Cosmos-Policy};
\end{tikzpicture}}
\captionof{figure}{\textbf{Visuomotor policies cannot see actuation shifts.} (a) A visual shift, such as a moved camera, changes the camera view. (b) An actuation shift does not: under altered arm gains, a $3\times$ stiffer controller or a 30\,N payload, the first frame and the command stay the same, but the arm moves differently (end-effector height, in simulation). (c) Under these shifts, the success of Cosmos-Policy on LIBERO-Spatial drops sharply; ``camera + gains'' combines the camera of (a) with the gains of (b). Our method, \ours, restores most of it.}
\label{fig:teaser}
\vspace{1.4em}

}]
{\renewcommand{\thefootnote}{}%
\footnotetext{$^{*}$Equal contribution.}%
\footnotetext{$^{\dagger}$Corresponding author: \href{mailto:j.liu4ai@gmail.com}{j.liu4ai@gmail.com}}}
\begin{abstract}
Visuomotor policies such as vision-language-action models (VLAs) and world action models (WAMs) are increasingly deployed on robots whose cameras and controllers differ from those used to collect their training data.
Such deployment introduces two kinds of shifts.
Visual shifts, such as a moved camera, change the observed images; actuation shifts, such as a retuned controller, latency or an added payload, change only how commanded actions translate into robot motion.
Invisible to the policy, actuation shifts sharply degrade its success, even under domain randomization.
In this work, we propose \ours, a test-time calibration method that identifies actuation shifts through active probing before task execution.
Specifically, before each episode, the robot executes a short probe that moves one arm axis at a time with known commands, so that gain, delay and drift leave separable traces.
A closed-form identifier estimates them from the response, and \ours compensates every subsequent command so that the robot executes the motion the policy intends.
The identifier requires no policy training and thus applies to any frozen visuomotor policy.
Since visual shifts remain visible to the policy, \ours further adapts the frozen policy with a removable low-rank adapter, addressing coupled visual and actuation shifts.
Experiments on LIBERO with a WAM and a VLA show that, without any policy training, the identifier of \ours recovers most of the success lost to actuation shifts and generalizes to shifts unseen during training (\eg, from 61.0\% to 89.7\% on LIBERO-Spatial with Cosmos-Policy).
\end{abstract}

\section{Introduction}
\label{sec:intro}

Visuomotor policies, which map camera observations and language instructions to robot actions, now achieve near-saturated success on standard manipulation benchmarks such as LIBERO~\cite{liu2023libero}.
This family includes vision-language-action models (VLAs)~\cite{kim2024openvla,kim2025oft,black2024pi0}, which fine-tune pretrained vision-language models to predict actions, and world action models (WAMs)~\cite{kim2026cosmospolicy,li2025uva,zhu2025uwm,ye2026dreamzero}, which jointly generate future frames and actions.
When such a policy is deployed on a different robot or at a different site, two aspects of the system can change: what the robot observes and how it executes commands.
A \emph{visual shift}, such as a displaced camera, as shown in \cref{fig:teaser}(a), alters the observed images.
An \emph{actuation shift}, such as a retuned controller, a control delay or an unmodeled payload, alters only the mapping from commanded actions to the resulting motion; as \cref{fig:teaser}(b) shows, the initial observation is unchanged.

An actuation shift is thus invisible to a policy that conditions only on its observations, and the resulting degradation is severe.
As shown in \cref{fig:teaser}(c), with the camera unchanged, tripling the controller stiffness reduces the success of the WAM Cosmos-Policy~\cite{kim2026cosmospolicy} on LIBERO-Spatial from 98\% to 15\%, and a 30\,N end-effector load reduces it to 9\%.
Domain randomization~\cite{tobin2017domain,peng2018sim} does not remove this limitation, since the observations still contain no information about the current actuation; on shifts absent from training, it even underperforms the released model.

Compensating an actuation shift thus requires evidence of the current dynamics.
Most adaptive methods infer the dynamics during execution, from the history or execution feedback of the current episode~\cite{kumar2021rma,lian2026reflective,jo2026jparc,lee2026selfcompensating,dey2026eavla,zhao2026apex} or from a previous failed trial~\cite{zhang2026selfadaptive}; this evidence arrives only after the policy has acted under the shift and stems from task motion not designed to reveal it.
Other methods move the arm randomly before the task, briefly~\cite{wang2026icwm} or for about 500 steps~\cite{lee2026space}; random excitation mixes the effects of axes, delays and loads, so that short probes are hard to read and long calibrations degrade nominal performance.
In both cases, the evidence is typically read by learned modules that generalize only to the dynamics seen in training.

Our key insight is that, since an actuation shift manifests only through motion, the robot should move before it acts, and move in a way that makes the shift easy to identify.
We therefore propose \ours (Probe, Estimate, Act), a test-time calibration method for frozen visuomotor policies.
Before each episode, the robot executes a short probe in free space that excites one arm axis at a time with zero-sum commands.
This design separates the unknowns: each axis's gain appears in its own response, a control delay shifts the response in time, and a constant external force leaves a net displacement.
From this response, a closed-form identifier, assembled from classical components~\cite{ljung1999system,swevers1997optimal,astrom1995adaptive} and calibrated once offline, estimates the per-axis gain, control delay and drift.
\ours then compensates every subsequent command, so that the robot executes the motion the policy intends.
This identifier requires no policy training and leaves the policy weights unchanged, so it applies to any policy with the same action interface.
For visual shifts, which the policy can observe, \ours further trains a lightweight low-rank adapter (LoRA)~\cite{hu2022lora} on the frozen weights, which lifts success under camera changes while largely preserving the policy's nominal performance.
We evaluate \ours on LIBERO with a WAM, Cosmos-Policy, and a VLA, OpenVLA-OFT~\cite{kim2025oft}, under coupled camera and gain shifts and under stiffness, payload and latency shifts absent from training.
Without any policy training, the identifier of \ours raises the success on LIBERO-Spatial under these unseen shifts from 61.0\% to 89.7\% on Cosmos-Policy and from 63.5\% to 89.0\% on OpenVLA-OFT.
With the adapter, \ours further raises the success of Cosmos-Policy on LIBERO-Spatial under coupled shifts from 42.8\% to 93.0\%.
\ours is thus policy-agnostic, needs only 30 steps and extends beyond training dynamics.
Our contributions are threefold:
\begin{itemize}[leftmargin=*,itemsep=1pt,topsep=2pt]
\item We distinguish visual and actuation shifts by whether a single frame reveals them, and show that actuation shifts, alone or coupled with camera shifts, severely degrade both WAMs and VLAs.
\item We propose \ours, a test-time calibration method that addresses both shifts: a 30-step probe read in closed form compensates actuation shifts without policy training, and a lightweight adapter handles visual shifts.
\item Extensive experiments on LIBERO with a WAM and a VLA demonstrate that \ours substantially improves the robustness of both policies to visual and actuation shifts, including dynamics never seen during training.
\end{itemize}

\definecolor{idc}{HTML}{E0662B}
\definecolor{lrc}{HTML}{2A78D6}
\definecolor{frz}{HTML}{8C8B86}
\definecolor{ink}{HTML}{3A3F47}
\begin{figure*}[t]
\centering
\resizebox{\textwidth}{!}{%
\begin{tikzpicture}[
  font=\footnotesize, line width=0.5pt,
  >={Stealth[length=1.6mm, width=1.3mm]},
  num/.style={circle, fill=#1, text=white, font=\footnotesize\bfseries, inner sep=0pt, minimum size=4.3mm},
  ttl/.style={font=\normalsize\bfseries, text=ink, anchor=west, inner sep=0pt},
  sub/.style={font=\scriptsize, text=black!50, anchor=west, inner sep=0pt},
  note/.style={font=\scriptsize, text=black!55},
  chip/.style={draw=#1!70, fill=#1!7, rounded corners=2pt, inner sep=3pt, font=\footnotesize, minimum height=4.6mm},
  flow/.style={->, line width=0.65pt, draw=ink!80},
  halo/.style={preaction={draw=white, line width=3.2pt, -}},
  tag/.style={fill=white, fill opacity=0.88, text opacity=1, inner sep=1.2pt, rounded corners=1pt, font=\scriptsize, anchor=north west},
]
\node[num=ink] (n1) at (0.25,5.6) {1};
\node[ttl] (t1) at ($(n1.east)+(0.13,0.03)$) {Probe};
\node[sub, anchor=base west] at ($(t1.base east)+(0.17,0)$) {before the episode, 30 steps};
\node[num=idc] (n2) at (5.85,5.6) {2};
\node[ttl] (t2) at ($(n2.east)+(0.13,0.03)$) {Estimate};
\node[sub, anchor=base west] at ($(t2.base east)+(0.17,0)$) {from the probe response};
\node[num=ink] (n3) at (11.4,5.6) {3};
\node[ttl] (t3) at ($(n3.east)+(0.13,0.03)$) {Act};
\node[sub, anchor=base west] at ($(t3.base east)+(0.17,0)$) {every step, base weights frozen};
\draw[black!12, line width=0.6pt] (5.45,1.45) -- (5.45,5.85);
\draw[black!12, line width=0.6pt] (11.0,1.45) -- (11.0,5.85);

\foreach \i/\a/\f in {0/{+x}/probe_0, 1/{+z}/probe_2, 2/{+r_z}/probe_5}{
  \node[anchor=south west, inner sep=0] (pf\i) at (0.2+\i*1.68,3.2) {\frameslot{\f.png}{1.55cm}{1.55cm}{frame while\\pushing $\a$}};
  \draw[black!25, line width=0.4pt, rounded corners=1.5pt] (pf\i.south west) rectangle (pf\i.north east);
  \node[tag] at ($(pf\i.north west)+(0.05,-0.05)$) {$\a$};
}
\def\xo{0.35}\def\dx{0.155}\def\amp{0.36}\def\yb{2.0}
\foreach \i in {0,2,4}{\fill[black!4] (\xo+2*\dx+4*\i*\dx,\yb-\amp-0.1) rectangle ++(4*\dx,2*\amp+0.2);}
\fill[black!9] (\xo+26*\dx,\yb-\amp-0.1) rectangle (\xo+30*\dx,\yb+\amp+0.1);
\node[font=\tiny, text=black!50] at (\xo+28*\dx,\yb) {grip};
\draw[black!25] (\xo,\yb) -- (\xo+30*\dx,\yb);
\foreach \i/\n in {0/x,1/y,2/z,3/r_x,4/r_y,5/r_z}{
  \pgfmathsetmacro{\xa}{\xo+(2+4*\i)*\dx}
  \pgfmathsetmacro{\xb}{\xa+2*\dx}
  \pgfmathsetmacro{\xc}{\xa+4*\dx}
  \draw[ink, line width=0.85pt, line join=round] (\xa,\yb) -- (\xa,\yb+\amp) -- (\xb,\yb+\amp) -- (\xb,\yb-\amp) -- (\xc,\yb-\amp) -- (\xc,\yb);
  \node[font=\scriptsize, text=black!60] at (\xb,\yb+\amp+0.25) {$\n$};
}
\draw[ink, line width=0.85pt] (\xo,\yb) -- (\xo+2*\dx,\yb);
\node[font=\scriptsize, rotate=90, text=black!55] at (0.1,\yb) {command};
\node[note, align=center] at (2.65,0.8) {one axis at a time, zero-sum: gain $\to$ own trace,\\delay $\to$ time shift, drift $\to$ net displacement};

\begin{scope}[shift={(6.05,3.6)}, x=0.42cm, y=1.35cm]
  \draw[black!40] (0,-0.62) -- (0,0.98);
  \draw[black!40] (0,0) -- (10.2,0);
  \node[font=\scriptsize, text=black!55, rotate=90, anchor=south] at (-0.12,0.5) {motion};
  \node[font=\scriptsize, text=black!55, anchor=north east] at (10.2,-0.03) {step};
  \fill[black!5] (2,0) -- (2,0.5) -- (4,0.5) -- (4,-0.5) -- (6,-0.5) -- (6,0) -- cycle;
  \draw[black!25, line width=0.6pt] (0,0) -- (2,0) -- (2,0.5) -- (4,0.5) -- (4,-0.5) -- (6,-0.5) -- (6,0);
  \draw[black!50, densely dashed, line width=0.8pt] plot[smooth, tension=0.45] coordinates
    {(0,0) (1,0.07) (2,0.081) (3,0.268) (4,0.439) (5,0.036) (6,-0.329) (7,-0.161) (8,-0.012) (9,-0.005)};
  \draw[idc, line width=1.2pt] plot[smooth, tension=0.45] coordinates
    {(0,0) (1,0.07) (2,0.081) (3,0.047) (4,0.373) (5,0.686) (6,0.094) (7,-0.571) (8,-0.335) (9,-0.054)};
  \fill[idc] (5,0.686) circle (0.9pt);
  \draw[idc!40, densely dotted] (4,0.439) -- (4,0.9);
  \draw[idc!40, densely dotted] (5,0.686) -- (5,0.9);
  \draw[<->, >={Stealth[length=1.2mm,width=1mm]}, idc!85!black, line width=0.55pt] (4,0.9) -- node[above, font=\scriptsize, inner sep=1.5pt] {$\hat d$} (5,0.9);
  \draw[idc!40, densely dotted] (5,0.686) -- (5.8,0.686);
  \draw[<->, >={Stealth[length=1.2mm,width=1mm]}, idc!85!black, line width=0.55pt] (5.8,0) -- node[right, font=\scriptsize, inner sep=1.5pt] {$\hat g_i$} (5.8,0.686);
  \begin{scope}[shift={(6.75,0.88)}, font=\scriptsize]
    \draw[black!25, line width=0.6pt] (0,0) -- (0.8,0); \node[anchor=west, inner sep=2pt, text=black!70] at (0.8,0) {command};
    \draw[black!50, densely dashed, line width=0.8pt] (0,-0.21) -- (0.8,-0.21); \node[anchor=west, inner sep=2pt, text=black!70] at (0.8,-0.21) {nominal $T_i$};
    \draw[idc, line width=1.2pt] (0,-0.42) -- (0.8,-0.42); \node[anchor=west, inner sep=2pt, text=black!70] at (0.8,-0.42) {response $y_i$};
  \end{scope}
\end{scope}
\node[chip=idc, anchor=west] (cg) at (5.85,2.42) {%
  $\hat g$\,\tikz[baseline=-0.5pt]{\foreach \h [count=\k] in {1.6,0.8,1.2,1.0,1.9,0.7}{\fill[idc, rounded corners=0.3pt] (\k*0.11,0) rectangle ++(0.075,\h*0.15);}\draw[black!35] (0.08,0.15)--(0.74,0.15);}};
\node[chip=idc, anchor=west] (cd) at ($(cg.east)+(0.13,0)$) {$\hat d=1$};
\node[chip=idc, anchor=west] (cb) at ($(cd.east)+(0.13,0)$) {$\hat b$ (drift)};
\node[sub, text=idc!85!black] at (5.9,1.88) {closed-form identifier, calibrated once, no training};
\draw[flow, rounded corners=3pt] (\xo+30*\dx+0.05,\yb) -- (5.45,\yb) |- (5.98,3.6);
\draw[flow, rounded corners=3pt] (5.45,\yb) |- (5.95,0.66);

\node[anchor=south west, inner sep=0] (obs) at (11.25,3.2) {\frameslot{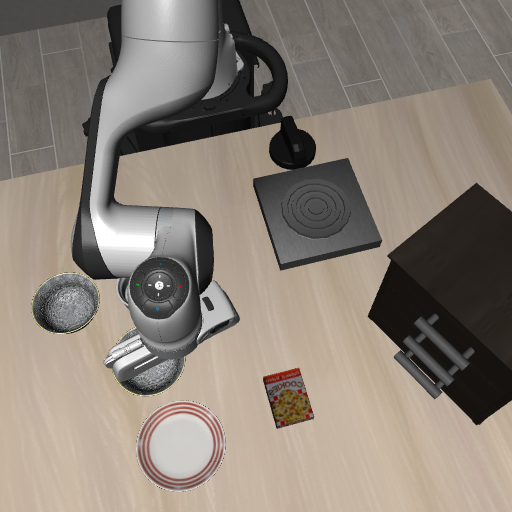}{1.55cm}{1.55cm}{task frame,\\moved camera}};
\draw[black!25, line width=0.4pt, rounded corners=1.5pt] (obs.south west) rectangle (obs.north east);
\node[tag] at ($(obs.north west)+(0.05,-0.05)$) {$o_t$};
\node[note, anchor=north] at ($(obs.south)+(0,-0.04)$) {moved camera};
\node[draw=frz!80, fill=frz!7, rounded corners=3pt, minimum width=2.1cm, minimum height=1.95cm, anchor=south west] (pol) at (13.15,2.95) {};
\node[font=\scriptsize, anchor=north, text=black!65, inner sep=2.5pt] at (pol.north) {policy (frozen)};
\foreach \k in {0,1,2}{
  \fill[frz!32, rounded corners=1pt] (13.35,3.13+\k*0.45) rectangle ++(1.08,0.3);
  \node[font=\tiny, text=black!75] at (13.89,3.28+\k*0.45) {$W_l$};
  \fill[lrc, rounded corners=1pt] (14.6,3.13+\k*0.45) rectangle ++(0.5,0.3);
}
\node[font=\tiny, text=white] at (14.85,4.18) {LoRA};
\draw[flow] (obs.east |- 0,3.97) -- (pol.west |- 0,3.97);
\node[anchor=west, inner sep=0] (chunk) at (15.6,3.92) {\tikz{\foreach \k in {0,...,4}{\draw[ink!70, fill=white, rounded corners=0.6pt] (\k*0.25,0) rectangle ++(0.2,0.2);}}};
\node[note, anchor=south] at ($(chunk.north)+(0,0.03)$) {chunk $a$};
\draw[flow] (pol.east |- chunk.west) -- (chunk.west);
\node[chip=idc, align=center, inner sep=3.5pt] (corr) at (16.78,2.45) {$\tilde a_i=\mathrm{clip}(a_i/\hat g_i+\hat b_i)$\\[-1pt]{\scriptsize hold $\hat d$ steps per chunk}};
\draw[flow] (chunk.south -| corr.north) ++(0,0) -- (corr.north);
\node[draw=black!40, fill=white, rounded corners=2pt, font=\scriptsize, inner sep=2.5pt] (env) at (16.78,1.56) {shifted env.\ $\rightarrow o_{t+1}$};
\draw[flow] (corr.south) -- (env.north);
\draw[->, idc, line width=0.8pt, halo] (cb.east) -- (corr.west |- cb.east);
\node[font=\scriptsize, text=idc!85!black, anchor=south] at (10.3,2.45) {$\hat g,\hat d,\hat b$};

\begin{scope}[on background layer]
  \draw[lrc!50, dashed, fill=lrc!3, rounded corners=4pt] (5.68,0.12) rectangle (18.62,1.22);
\end{scope}
\node[draw=lrc!80, fill=lrc!10, rounded corners=2pt, inner sep=3pt, font=\scriptsize, anchor=west] (enc) at (5.95,0.66) {probe encoder};
\node[anchor=west, inner sep=0] (sv) at ($(enc.east)+(0.35,0)$) {\tikz{\foreach \v [count=\k] in {80,35,60,15,90,45,25,70,10,55,30,85,20,50}{\fill[lrc!\v, rounded corners=0.3pt] (\k*0.13,0) rectangle ++(0.105,0.26);}}};
\node[font=\scriptsize, text=lrc!85!black, anchor=south] at (sv.north) {spectrum $s(z)$};
\draw[->, lrc, line width=0.65pt] (enc.east) -- (sv.west);
\begin{scope}[shift={(13.6,0.42)}, scale=0.8, every node/.style={transform shape}]
  \fill[frz!32, rounded corners=1pt] (0,0) rectangle (0.75,0.75); \node[font=\scriptsize] at (0.375,0.375) {$W_l$};
  \node[font=\footnotesize] at (0.92,0.375) {$+$};
  \fill[lrc, rounded corners=0.5pt] (1.08,0) rectangle (1.26,0.75); \node[font=\scriptsize, below, inner sep=1pt] at (1.17,0) {$B_l$};
  \draw[lrc] (1.33,0.08) rectangle (1.92,0.67);
  \foreach \v [count=\k] in {85,40,65,20}{\fill[lrc!\v] (1.33+\k*0.12-0.09,0.67-\k*0.12-0.02) rectangle ++(0.11,0.11);}
  \fill[lrc, rounded corners=0.5pt] (1.98,0.29) rectangle (2.53,0.46); \node[font=\scriptsize, above, inner sep=1pt] at (2.25,0.46) {$A_l$};
\end{scope}
\coordinate (adtop) at (14.85,1.0);
\draw[->, lrc, line width=0.65pt] (sv.east) -- node[above, font=\scriptsize, text=lrc!85!black, inner sep=1.5pt] {$s_l(z)$ per layer} (13.5,0.66);
\draw[->, lrc, line width=0.8pt, halo] (adtop) -- (14.85,2.95);
\node[font=\scriptsize, text=lrc!85!black, align=left, anchor=west, inner sep=1pt] at (15.9,0.67) {\textbf{LoRA adapter}\\for visual shifts\\(the only learned part)};
\end{tikzpicture}}
\caption{\textbf{Overview of PEA.} Actuation shifts are invisible in images but revealed by motion, so PEA moves before it acts. (1)~\textbf{Probe}: before each episode, the robot executes a 30-step probe that pushes one arm axis at a time with zero-sum commands, so that each gain, the delay and a constant drift leave separable traces. (2)~\textbf{Estimate}: a closed-form identifier (orange), calibrated once offline and requiring no policy training, reads per-axis gain $\hat g$, delay $\hat d$ and drift correction $\hat b$ from the response (curves: logged $x$ motion under gain 1.6 and a 1-step delay). (3)~\textbf{Act}: every command of the frozen policy (gray) is rescaled and offset, and the arm holds for $\hat d$ steps after each chunk; a lightweight LoRA adapter (blue) on the frozen weights, conditioned on the probe, handles visual shifts such as a moved camera.}
\label{fig:method}
\end{figure*}

\section{Related Work}
\label{sec:related}

\paragraph{Robustness of VLAs and WAMs.}
Most robustness benchmarks for VLAs and WAMs perturb what a policy observes or reads~\cite{fei2025liberoplus,zhou2025liberopro,xiao2026vlagauge,sawada2026liberoctrl}, or inject action noise~\cite{guo2026robustvla}.
WAMs~\cite{li2025uva,zhu2025uwm,hu2025vpp,kim2026cosmospolicy,ye2026dreamzero,chen2026lawam} are reported to be more robust than VLAs to such visual perturbations~\cite{zhang2026wamrobust}, and dedicated remedies target camera changes~\cite{heo2026anycamvla,park2026gsvla,li2025vlageneralizable,kang2026dart}.
More generally, a frozen policy can be adapted through prompt tokens~\cite{li2021prefix,lester2021prompt}, feature-wise modulation~\cite{perez2018film} or low-rank updates~\cite{hu2022lora}, possibly generated by a hypernetwork~\cite{ha2017hypernetworks,mahabadi2021hyperformer,phang2023hypertuning,charakorn2025t2l,xiong2024hyperlora,bianchi2026wizard,palafox2026claw}; \ours uses such an adapter for visual shifts.
Far fewer works vary the robot itself: LIBERO-RIG~\cite{cui2026liberorig} changes actuator gain and control rate, where Cosmos-Policy is among the most affected models, and joint-level faults~\cite{jo2026jparc} and execution errors~\cite{lee2026selfcompensating} also degrade VLAs.
We distinguish shifts by whether a single frame reveals them, extend actuation shifts to stiffness, payload and latency, couple them with camera shifts, and correct them rather than only measure them.

\paragraph{Adaptation to dynamics shifts.}
Adaptive policies update their weights or prompts at test time~\cite{liu2026ttvla,zhang2026tttvla,zang2026pdf,lee2026salt,jiang2026robottt,li2026recastvla,yuan2026adaworldpolicy,zhou2026ewam}, infer the dynamics from the interaction history~\cite{lian2026reflective,lyu2025dywa,zhang2026selfadaptive,shi2026vlrla,yu2017preparing,kumar2021rma,qi2022hora,rakelly2019pearl,nagabandi2019learning}, or are trained on randomized domains~\cite{tobin2017domain,peng2018sim}, and thus cover only the dynamics seen in training.
Closest to our problem, execution-aware policies~\cite{jo2026jparc,lee2026selfcompensating,dey2026eavla} and online tracking compensation~\cite{zhao2026apex} exploit the gap between commanded and executed motion during the episode.
Closest in the evidence used, ICWM~\cite{wang2026icwm} keeps clips of random pre-task motion in a VLA's context, and SPACE~\cite{lee2026space} fits an affine command adapter on about 500 random calibration steps.
Grounded action transformation~\cite{hanna2017grounded} likewise rewrites the actions of a fixed policy, and real-time chunking~\cite{black2025rtc} compensates inference delay.
Unlike these methods, \ours gathers its evidence before the task with a short probe designed for identification, reads it with an explicit model rather than a learned module, and keeps the policy frozen.

\paragraph{System identification and active perception.}
Identifying a robot's dynamics from designed excitation is a long-standing topic in control: optimized excitation makes the parameters of manipulator models identifiable~\cite{swevers1997optimal}, least-squares and kernel methods estimate them from input-output data~\cite{ljung1999system,pillonetto2014kernel}, and certainty-equivalence control acts on the estimates as if they were exact~\cite{astrom1995adaptive}; our identifier assembles these components.
In robot learning, environment-probing policies~\cite{zhou2019epi}, active exploration for system identification~\cite{memmel2024asid} and learned in-context exploration~\cite{poddar2026asteroid} choose actions that reveal physical parameters, and interactive perception more broadly acts in order to perceive~\cite{bohg2017interactive}.
Hand-eye calibration~\cite{tsai1989handeye} also moves the robot in view of the camera, but to recover geometry rather than dynamics.
Whereas these works learn the exploration or the estimator, or calibrate a robot once, \ours calibrates a frozen, image-conditioned and chunked VLA or WAM within a budget of 30 steps per episode, a setting in which we find that the probe's design largely determines the identification error.

\section{Preliminaries}
\label{sec:method:prelim}

\subsection{Visuomotor policies}
A pretrained policy $\pi$ maps an observation $o_t$ (camera images and proprioception) and a language instruction $\ell$ to a chunk of $H$ actions $(a_t,\dots,a_{t+H-1})=\pi(o_t,\ell)$, which is executed open loop before the next observation is taken.
Each action $a_t\in\mathbb{R}^7$ consists of six end-effector translation and rotation deltas $a_{t,1},\dots,a_{t,6}$ for an operational-space controller (OSC) and a gripper command $a_{t,7}$.
We study two such policies: the WAM Cosmos-Policy~\cite{kim2026cosmospolicy}, built on the Cosmos-Predict2-2B video model~\cite{nvidia2025cosmos} with T5~\cite{raffel2020t5} instruction embeddings, and the VLA OpenVLA-OFT~\cite{kim2025oft}.

\subsection{Visual and actuation shifts}
\label{sec:prelim:shifts}
We separate the system the policy acts in into an \emph{observation map}, which renders the scene into $o_t$ through camera parameters $\psi$, and an \emph{execution map}, which turns a command $a_t$ into the end-effector motion $\Delta x_t$ through actuation parameters $\theta$ such as controller gains, stiffness, latency and external forces.
A \emph{visual shift} changes $\psi$; an \emph{actuation shift} changes $\theta$.
The two differ in what the policy can perceive.
A visual shift alters every frame, including the first.
An actuation shift leaves the initial observation $o_0$ unchanged, because $o_0$ depends on $\theta$ only through motion that has not happened yet; it becomes observable only once the robot moves.
A policy that acts from $o_0$ therefore commits to its first action chunk blind to the actuation that executes it.

\subsection{Problem formulation}
\label{sec:method:setting}
Let $(\psi_0,\theta_0)$ denote the camera and actuation parameters under which $\pi$ was trained.
At test time, each episode runs under unknown parameters $(\psi,\theta)$ that are constant within the episode and may differ from $(\psi_0,\theta_0)$ in $\psi$, in $\theta$, or in both.
Our goal is to maximize the task success of $\pi$ under such shifts, subject to three deployment constraints.
(C1)~\emph{Frozen policy}: the pretrained weights of $\pi$ remain unchanged; only an adapter attached to them may be trained.
(C2)~\emph{Bounded interaction}: before each episode, the robot may execute at most $K{=}30$ steps in free space.
(C3)~\emph{No test-time tuning}: no component is tuned on test shifts; offline training uses only episodes from a training family of shifts, demonstrated by a privileged \teach (\cref{sec:method:train,sec:setup}).

\section{Methodology}
\label{sec:method}

In this section, we introduce \ours, a test-time calibration method that identifies actuation shifts with a short probe and compensates every command accordingly, while a lightweight adapter on the frozen policy handles visual shifts.
Specifically, as shown in \cref{fig:method}, the actuation calibration proceeds in three steps.
First, before each episode, the robot executes a short, fixed \emph{probe} in free space and records its own motion.
Then, a closed-form identifier \emph{estimates} the per-axis gain, control delay and drift from this motion.
Finally, \ours compensates every command of the policy for the identified actuation, so that the robot \emph{acts} as the policy intends.
In addition, a LoRA adapter conditioned on the same probe adapts the frozen policy to visual shifts.

\subsection{Actuation model}
\label{sec:method:model}
To identify an actuation shift within a few steps, \ours approximates the execution map of each arm axis $i\in\{1,\dots,6\}$ with a low-dimensional model,
\begin{equation}
\Delta x_{t,i}\approx k_i\,g_i\,a_{t-\delta,i}+\beta_i ,
\label{eq:model}
\end{equation}
where $k_i$ is the nominal motion per unit command.
The unknowns, which summarize $\theta$, are a per-axis action gain $g_i$, a control delay $\delta$ shared by all axes and a constant drift $\beta_i$ caused by an external force; the nominal actuation corresponds to $g_i{=}1$, $\delta{=}0$ and $\beta_i{=}0$.
The model is deliberately simple: it has few enough unknowns to be identified within 30 steps, yet it captures the first-order effects of a retuned controller, a latency and a payload.
A stiffness change enters it only approximately, as an effective gain and lag, whereas cross-axis coupling and time-varying loads lie outside it (\cref{sec:limitations}).
Given the parameters, gain and drift can be inverted exactly: the command $\tilde a_{t,i}=a_{t,i}/g_i+b_i$ with $b_i=-\beta_i/(k_ig_i)$ makes \cref{eq:model} return the nominal motion $k_ia_{t-\delta,i}$.
The delay, in contrast, cannot be undone by issuing commands early, since the policy decides each chunk only after observing; \cref{sec:method:act} handles it with a hold rule.

\subsection{Probe design}
\label{sec:method:probe}
The probe is designed so that each unknown of \cref{eq:model} leaves a separate trace that a closed-form estimator can read.
Its $K{=}30$ steps start with two no-op steps that let the arm settle.
Each arm axis in turn then receives the doublet $+c,+c,-c,-c$ with $c{=}0.5$, while all other axes stay at zero, and four final steps close and open the empty gripper.
We measure $\Delta x_t$ as the end effector's per-step position change $100\,(p_{t+1}-p_t)$ and rotation change $10\,\mathrm{rotvec}(R_{t+1}R_t^{\top})$, scaled to a similar range.
For axis $i$, the response $y_i\in\mathbb{R}^L$ ($L{=}7$) is the channel $\Delta x_{t,i}$ over the four push steps of axis $i$ and the next three, which capture a delay of up to three steps.
The zero-sum doublets and the one-axis-at-a-time excitation separate the unknowns.

\begin{proposition}
\label{prop:sep}
Assume \cref{eq:model} holds without noise, $\delta\le3$, and let $T_i$ be the nominal doublet response of axis $i$. Then
(i)~the net displacement of the probe, $D_i=\sum_{t=1}^{K}\Delta x_{t,i}$, equals $K\beta_i$ for any gains and delay; and
(ii)~fitting $y_i\approx\alpha_i\,S_dT_i$ by least squares, where $(S_dT_i)_t=T_{i,t-d}$ (zero for $t\le d$) delays $T_i$ by $d$ steps, gives $\alpha_i=g_i$ at $d=\delta$ for any drift, and the explained energy is maximal only at $d=\delta$.
\end{proposition}
\noindent\emph{Proof sketch.}
(i)~The commands of each axis sum to zero and, with $\delta\le3$, are all executed within the probe, so the gain term of \cref{eq:model} sums to zero and only $K\beta_i$ remains.
(ii)~$y_i=g_iS_\delta T_i+\beta_i\mathbf{1}$, and the zero-sum doublet gives $\langle\mathbf{1},S_dT_i\rangle=0$, so the drift does not bias the fit; by Cauchy--Schwarz, $\langle y_i,S_dT_i\rangle^2/\lVert S_dT_i\rVert^2\le g_i^2\lVert S_\delta T_i\rVert^2$, with equality only when $S_dT_i\parallel S_\delta T_i$, \ie, $d=\delta$.\hfill$\square$

\noindent The zero-sum doublet thus decouples drift from gain and delay, while exciting one axis at a time keeps each response free of the others and limits the coupling and nonlinearity that \cref{eq:model} omits (\cref{sec:res:evidence}).
After the probe, our main evaluation resets the simulator; alternatively, a servo return ends within 2\,mm of the start and moves no object (\cref{sec:supp:return}).

\subsection{Closed-form identification}
\label{sec:method:id}
Following \cref{prop:sep}(ii), the identifier fits each response as a scaled, possibly delayed copy of its nominal template, $y_i\approx\alpha_i\,S_dT_i$, where $T_i$ is estimated as the mean of $y_i$ over training probes with nominal actuation.
For each candidate delay $d\in\{0,1,2,3\}$, the least-squares amplitude and the response energy explained over all arm axes are
\begin{equation}
\alpha_i(d)=\tfrac{\langle y_i,S_dT_i\rangle}{\lVert S_dT_i\rVert^2},\qquad
E(d)=\textstyle\sum_{i=1}^{6}\tfrac{\langle y_i,S_dT_i\rangle^2}{\lVert S_dT_i\rVert^2}.
\label{eq:amp}
\end{equation}
The identifier selects $\hat d=\arg\max_d E(d)$, pooling all axes because they share one delay.
The amplitude $\alpha_i(\hat d)$ equals $g_i$ only if the controller is linear, as \cref{eq:model} assumes; the real controller responds nonlinearly to the amplitude, most on rotation axes (\cref{sec:supp:superpos}).
The identifier therefore sets $\hat g_i=\phi_i\big(\alpha_i(\hat d)\big)$, where $\phi_i$ is a per-axis quadratic calibration fitted once by least squares on 800 training probes, the identifier's only use of privileged gains; without it, the mean gain error on unseen in-range probes rises from 3.2\% to 6.4\% (\cref{sec:supp:calcheck}).

For the drift, \cref{prop:sep}(i) suggests reading the net displacement $D_i$.
On the real controller, $D_i$ is not exactly zero without a force, but it stays near its mean $\mu_i$ over force-free training probes under any gain or delay.
Since external forces act on position, a drift test flags a translation axis $i\in\{1,2,3\}$ when $|D_i-\mu_i|>3\sigma_i$, with $\sigma_i$ the standard deviation over the same probes, estimates the drift as $\hat\beta_i=(D_i-\mu_i)/K$ and sets
\begin{equation}
\hat b_i=\operatorname{clip}\!\big(-\hat\beta_i/(k_i\,\hat g_i),\,-0.5,\,0.5\big),
\label{eq:bias}
\end{equation}
the estimate of $b_i$ in \cref{sec:method:model}; all other axes get $\hat b_i=0$.

\subsection{Command compensation}
\label{sec:method:act}
At every step, \ours inverts gain and drift with the estimates $\hat g_i,\hat b_i$ and leaves $a_{t,7}$ unchanged:
\begin{equation}
\tilde a_{t,i}=\operatorname{clip}\!\big(a_{t,i}/\hat g_i+\hat b_i,\,-1,\,1\big),\quad i=1,\dots,6.
\label{eq:act}
\end{equation}
Under a delay $\delta$, the last $\delta$ commands of a chunk would still be executing when the next observation is taken, so the policy would plan from a state that its own commands are about to change.
The \emph{hold rule} therefore sends zero arm commands for $\hat d$ steps after each chunk, so that every chunk completes before the next observation; held steps count against the episode's step budget.
Since probing, identification and compensation only read the robot's motion and rewrite commands, they are policy-agnostic and serve both policies unchanged (\cref{sec:supp:impl}).
We denote this part of \ours, which requires no policy training, by \pid.

\subsection{Adapter for visual shifts}
\label{sec:method:lora}
\label{sec:method:train}
Visual shifts, such as a moved camera, change the images the policy conditions on and lie outside \cref{eq:model}; unlike actuation shifts, they are observable, so the policy can learn to handle them.
\ours adapts the frozen policy with a low-rank adapter (LoRA)~\cite{hu2022lora} on all 168 linear layers of the Cosmos-Policy diffusion transformer (224 layers of the Llama backbone for OpenVLA-OFT).
Each layer becomes
\begin{equation}
W_l(z)=W_l+B_l\,\mathrm{diag}\big(s_l(z)\big)A_l ,
\label{eq:lora}
\end{equation}
with $W_l$ frozen, rank-14 factors $A_l,B_l$ shared across episodes, and a per-episode spectrum $s(z)=\{s_l(z)\}_l$ generated from an embedding $z$ that a probe encoder computes from the probe's motion and frames (\cref{sec:supp:impl}).
The adapter adds 16.9M parameters besides its encoder, and the base weights stay bit-identical, as (C1) requires.

Training data come from a privileged \teach on the training family of shifts (\cref{sec:setup}): the released model, which sees the nominal camera view and executes its arm actions $\bar a$ divided elementwise by the true gains $g=(g_1,\dots,g_6)$.
The adapter is trained with the policy's own objective on the \emph{intended label} $\bar a$, \ie, the action in nominal space, rather than on the executed label $\bar a/g$: since the identifier already compensates the actuation at test time, the latter would compensate the gain twice (\cref{tab:abl}).

\begin{table*}[t]
\centering\vspace{-2pt}
\small
\setlength{\tabcolsep}{3.5pt}
\renewcommand{\arraystretch}{0.92}\setlength{\aboverulesep}{0.25ex}\setlength{\belowrulesep}{0.15ex}
\newcolumntype{C}{>{\centering\arraybackslash}p{0.95cm}}
\newcommand{\gl}[1]{{\scriptsize\color{black!55}#1}}%
\newcommand{\grp}[1]{\multicolumn{10}{l}{\cellcolor{black!6}\textit{#1}}}%
\begin{tabular}{l c CC CC CC CC}
\toprule
\multirow{2}{*}{Method} & \multirow{2}{*}{Pre-task} & \multicolumn{2}{c}{Nominal} & \multicolumn{2}{c}{Seen types} & \multicolumn{2}{c}{Unseen types} & \multicolumn{2}{c}{New dynamics} \\
\cmidrule(lr){3-4} \cmidrule(lr){5-6} \cmidrule(lr){7-8} \cmidrule(lr){9-10}
 & & Spatial & Goal & Spatial & Goal & Spatial & Goal & Spatial & Goal \\
\midrule
\grp{Frozen policy} \\
\raw & 0 & 98\rlap{$^\dagger$} & 98\rlap{$^\dagger$} & 42.8 & 45.8 & 40.8 & 55.3 & 61.0 & 69.7 \\
SPACE-style~\cite{lee2026space} \gl{calibration} & 500 & 74 & 65 & 44.0 & 49.0 & 33.3 & 38.3 & 71.8 & 61.5 \\
ICWM-style~\cite{wang2026icwm} \gl{random probe} & 30 & 96 & 89 & 60.5 & 67.7 & 47.5 & 44.8 & 66.5 & 65.5 \\
\gid \ \gl{gain} & 30 & 99 & 98 & 77.3 & 82.3 & 70.3 & 70.3 & 38.0 & 32.2 \\
\pid{} w/o drift \gl{gain, delay} & 30 & 98 & 97 & 75.3 & 83.0 & 69.5 & 68.0 & 78.0 & 71.5 \\
\rowcolor{cvprblue!9} \textbf{\pid} \ \gl{gain, delay, drift} & 30 & 100 & 97 & 76.3 & 82.3 & 68.8 & 68.3 & 89.7 & \textbf{84.0} \\
\pid{} + prefix adapter \gl{frozen base} & 30 & 85 & 58 & 63.0 & 39.5 & 63.5 & 50.0 & 74.2 & 56.0 \\
\rowcolor{cvprblue!9} \textbf{\oursfull}\ \gl{+ LoRA adapter, frozen base} & 30 & 99 & 98 & \textbf{93.0} & \textbf{91.5} & \textbf{87.8} & \textbf{80.0} & \textbf{90.3} & 79.7 \\
\midrule
\grp{Fine-tuned policy behind an estimator} \\
FT + SPACE-style & 500 & 79 & 65 & 72.0 & 68.3 & 65.3 & 61.5 & 76.3 & 59.7 \\
FT + ICWM-style & 30 & 97 & 94 & 78.5 & 81.0 & 72.5 & 65.5 & 64.5 & 64.5 \\
FT + ARX \gl{delay, RLS} & 500 & 95 & 83 & 76.3 & 61.8 & 85.0 & 70.8 & 86.8 & 66.0 \\
FT + \gid & 30 & 99 & 95 & \textbf{94.0} & 95.8 & \textbf{91.0} & \textbf{86.5} & 36.8 & 32.5 \\
\oursfull-FT \gl{joint fine-tuning} & 30 & 99 & 98 & 93.0 & \textbf{96.2} & 90.5 & \textbf{86.5} & \textbf{90.5} & \textbf{82.8} \\
\bottomrule
\end{tabular}
\caption{\textbf{Comparison with baselines on Cosmos-Policy.} Success rates (\%) on LIBERO-Spatial/Goal under nominal conditions and the three shift types of \cref{sec:setup}. Pre-task: steps executed before each episode. Bold marks the best result per column within each group, excluding nominal; $^\dagger$: release protocol (500 episodes). Per-configuration results are in \cref{sec:supp:cells}.}
\label{tab:main}
\end{table*}

\section{Experiments}
\label{sec:experiments}
\label{sec:results}

We evaluate whether \ours corrects actuation shifts without policy training, whether this extends to dynamics absent from training, and whether its adapter handles coupled visual and actuation shifts across policies; we then analyze the design choices behind these results.

\subsection{Experimental setup}
\label{sec:setup}

\paragraph{Benchmark and shifts.}
We evaluate on the LIBERO~\cite{liu2023libero} benchmark with test shifts fixed per configuration, in three groups.
\emph{Seen types} couple per-axis action gains, the gripper force limit and a camera rotation, drawn from the training family ($g_i\sim\mathcal{U}[0.7,2.0]$, azimuth within $\pm40^\circ$, elevation within $25^\circ$).
\emph{Unseen types} add a $1.5\times$ zoom-out, object masses $\times100$ or a narrower field of view.
\emph{New dynamics} each change one factor that training never varies: OSC stiffness $\times0.3$ or $\times3$, a 15 or 30\,N downward force on the end effector, or a 1- or 2-step command latency.
\Cref{tab:supp:cfg} lists all test configurations, and the appendix further reports seen-type couplings with added latency or payload, shifts that violate \cref{eq:model} such as cross-axis mixing, and the hardware-setup benchmark LIBERO-RIG~\cite{cui2026liberorig}.

\paragraph{Policies and baselines.}
We adopt the WAM Cosmos-Policy~\cite{kim2026cosmospolicy} and the released four-suite checkpoint of the VLA OpenVLA-OFT~\cite{kim2025oft} as the visuomotor policies.
Adapters are trained once per suite pair (LIBERO-Spatial with LIBERO-Goal, LIBERO-Object with LIBERO-Long) on successful \teach episodes whose initial states are disjoint from the test states.
\Cref{tab:main} compares two groups.
The first keeps the released policy frozen and places an estimator in front of it: besides \pid and \oursfull, \gid reads only the gain, as each axis's mean probe response relative to nominal; our re-implementations of the evidence of ICWM~\cite{wang2026icwm} (a random 30-step probe) and SPACE~\cite{lee2026space} (500 calibration steps with online LMS) are marked ``-style''; and a prefix-token adapter replaces our LoRA.
The second serves as a reference that fine-tunes the policy: the base is fine-tuned on intended labels without probe input (FT) and paired with these estimators or an ARX identifier with RLS~\cite{ljung1999system}, and \oursfull-FT trains our adapter jointly with the base.
SPACE-style and ARX also calibrate for 500 steps rather than 30.

\paragraph{Evaluation metric.}
We report the task success rate (\%).
Each test configuration is evaluated on 100 episodes per suite (10 tasks $\times$ 10 initial states), and each table column averages the configurations of one shift type.
Unless noted, results are for Cosmos-Policy; \cref{sec:supp:setup} gives training details and statistical tests.

\begin{table}[t]
\centering
\small
\setlength{\tabcolsep}{2.4pt}
\renewcommand{\arraystretch}{0.95}
\newcommand{\B}[1]{\textbf{#1}}
\begin{tabular}{l ccc ccc ccc}
\toprule
\multirow{2}{*}{Suite} & \multicolumn{3}{c}{Seen types} & \multicolumn{3}{c}{Unseen types} & \multicolumn{3}{c}{New dynamics} \\
\cmidrule(lr){2-4} \cmidrule(lr){5-7} \cmidrule(lr){8-10}
 & Orig. & ID & \oursfull & Orig. & ID & \oursfull & Orig. & ID & \oursfull \\
\midrule
Spatial & 35.0 & 48.5 & \B{83.7} & 49.0 & 61.5 & \B{74.0} & 63.5 & \B{89.0} & 84.2 \\
Goal    & 35.3 & 55.2 & \B{84.2} & 43.5 & 54.0 & \B{68.0} & 64.8 & \B{82.3} & 79.8 \\
Object  & 78.7 & 91.8 & \B{94.2} & 40.0 & 48.3 & \B{53.8} & 62.2 & \B{84.8} & 82.8 \\
Long    & 22.2 & 36.8 & \B{70.2} & 13.3 & 19.5 & \B{21.8} & 55.5 & \B{77.3} & 69.3 \\
\bottomrule
\end{tabular}
\caption{\textbf{Results on a second policy, OpenVLA-OFT.} Success rates (\%) on the four LIBERO suites under the three shift types of \cref{sec:setup}. Orig.: released policy; ID: \pid, applied unchanged from Cosmos-Policy; \oursfull: \pid with the adapter. Bold marks the best result per suite and shift type. }
\label{tab:gen}
\end{table}

\begin{table*}[!t]
\centering
\small
\setlength{\tabcolsep}{4pt}
\renewcommand{\arraystretch}{0.92}
\newcommand{\y}{\checkmark}
\newcommand{\n}{--}
\begin{tabular}{l ccc cc cc cc cc}
\toprule
\multirow{2}{*}{Variant} & \multicolumn{3}{c}{Identifier reads} & \multicolumn{2}{c}{Adapter} & \multicolumn{2}{c}{Seen types} & \multicolumn{2}{c}{Unseen types} & \multicolumn{2}{c}{New dynamics} \\
\cmidrule(lr){2-4} \cmidrule(lr){5-6} \cmidrule(lr){7-8} \cmidrule(lr){9-10} \cmidrule(lr){11-12}
 & gain & delay & drift & LoRA & probe input & Spatial & Goal & Spatial & Goal & Spatial & Goal \\
\midrule
\raw & \n & \n & \n & \n & \n & 42.8 & 45.8 & 40.8 & 55.3 & 61.0 & 69.7 \\
\gid & \y & \n & \n & \n & \n & 77.3 & 82.3 & 70.3 & 70.3 & 38.0 & 32.2 \\
\pid{} w/o drift & \y & \y & \n & \n & \n & 75.3 & 83.0 & 69.5 & 68.0 & 78.0 & 71.5 \\
\pid & \y & \y & \y & \n & \n & 76.3 & 82.3 & 68.8 & 68.3 & 89.7 & 84.0 \\
static LoRA & \y & \y & \y & \y & zeroed & 90.8 & 90.3 & 89.3 & 80.0 & 92.0 & 78.7 \\
\rowcolor{cvprblue!9} \textbf{\oursfull} & \y & \y & \y & \y & frames + motion & 93.0 & 91.5 & 87.8 & 80.0 & 90.3 & 79.7 \\
\quad frames zeroed at test & \y & \y & \y & \y & motion & 94.5 & 90.5 & 88.3 & 79.8 & 89.5 & 81.2 \\
\quad motion zeroed at test & \y & \y & \y & \y & frames & 73.5 & 78.7 & 70.0 & 68.5 & 90.0 & 83.7 \\
\bottomrule
\end{tabular}
\caption{\textbf{Ablation study on Cosmos-Policy.} Starting from the released policy (\raw), the components of \oursfull are added one at a time, with the policy weights frozen throughout; success rates (\%) on LIBERO-Spatial/Goal. \emph{Identifier reads}: the actuation parameters estimated from the probe and compensated. \emph{Adapter}: whether the LoRA adapter is used and which probe signals it receives. The static LoRA is the same adapter trained with its probe input zeroed; the two indented rows zero one probe input of the trained adapter at test time. --: not used.}
\label{tab:comp}
\end{table*}

\subsection{Main results}
\label{sec:res:main}
\label{sec:res:oft}

\Cref{tab:main} compares \ours with all baselines on Cosmos-Policy (LIBERO-Spatial/Goal), and \cref{tab:gen} reports \ours on OpenVLA-OFT across all four LIBERO suites.
Both tables first confirm the problem raised in \cref{sec:intro}: on LIBERO-Spatial, the success rate of Cosmos-Policy drops from 98\% under nominal conditions to 42.8\% on seen types, and that of OpenVLA-OFT from 98\% to 35.0\%.
\ours recovers most of this loss on both policies with their weights frozen.

\paragraph{The identifier corrects actuation shifts without policy training.}
Reading only the probe, \pid raises the success rate of Cosmos-Policy on LIBERO-Spatial from 42.8\% to 76.3\% on seen types and from 40.8\% to 68.8\% on unseen types, while its nominal success stays at 100\%.
On the seen-type coupling dominated by gain shifts, it restores 98\% on both LIBERO-Spatial and LIBERO-Goal: once the actuation is compensated, the frozen policy acts as on the nominal robot.
The remaining gap comes from the camera, which the identifier by design does not read.
It is largest on OpenVLA-OFT, which was never trained for camera changes: the same identifier, unchanged, lifts its seen types on LIBERO-Spatial only from 35.0\% to 48.5\%.

\paragraph{The adapter handles the visual shifts that remain.}
Where the camera changes most, \pid reaches only 48\% on LIBERO-Spatial under a seen-type coupling with 23$^\circ$ camera elevation and 60\% under an unseen-type $1.5\times$ zoom-out; the adapter lifts these two configurations to 95\% and 93\%.
Overall, on LIBERO-Spatial, \oursfull reaches 93.0\% on seen and 87.8\% on unseen types with Cosmos-Policy, and raises the seen types of OpenVLA-OFT from 48.5\% to 83.7\%.
Since visual shifts are observable, an adapter on frozen weights suffices: training it jointly with the base (\oursfull-FT) adds at most 2.7 points on LIBERO-Spatial and 6.5 points on LIBERO-Goal across seen and unseen types.
The two components are complementary and therefore compose: on seen-type couplings that also add latency or payload, \raw, \pid and \oursfull reach 28.5\%, 69.3\% and 87.5\% on LIBERO-Spatial.

\paragraph{Comparison with baselines.}
Among systems that keep the released weights, \pid outperforms the ICWM-style random probe of the same length (60.5\% on LIBERO-Spatial seen types), which excites all axes at once and, as analyzed in \cref{sec:res:evidence}, identifies the gains less accurately.
The SPACE-style calibration runs 500 steps yet reaches only 44.0\% on LIBERO-Spatial seen types and lowers nominal success to 74\%.
On LIBERO-Spatial new dynamics, both baselines also stay below \pid (71.8\% and 66.5\% vs.\ 89.7\%).
Fine-tuning the base behind an estimator does not replace the identifier either: FT + \gid matches or exceeds \oursfull on seen and unseen types, which contain no latency or force, but collapses to 36.8\% on LIBERO-Spatial new dynamics, where reading the gain alone fails, and FT + ARX stays below \oursfull on every column.

\begin{figure}[t]
\centering
\includegraphics[width=\linewidth]{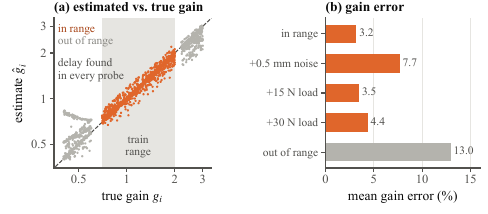}
\caption{\textbf{Identification accuracy on unseen probes.} (a)~Estimated vs.\ true per-axis gain on 360 probes not used for calibration: 120 in the training range, 120 with a 15 or 30\,N load and 120 outside the range. (b)~Mean gain error; the noise bar adds 0.5\,mm Gaussian noise per step. The delay is exact for all probes.}
\label{fig:idacc}
\end{figure}

\subsection{Generalization to new dynamics}
\label{sec:res:dyn}

Since the identifier reads an explicit model, it extends to actuation shifts that training never contained.
Without policy training, \pid raises the new-dynamics success on LIBERO-Spatial from 61.0\% to 89.7\% for Cosmos-Policy and from 63.5\% to 89.0\% for OpenVLA-OFT.
These cells lie inside \cref{eq:model} by construction, and \cref{fig:idacc} shows that the estimates remain accurate on probes the calibration never saw: the gain error is 3.2\% in the training range and 4.4\% under a 30\,N load, and the delay is exact for all probes.

\paragraph{Latency.}
A delay of one or two control steps barely affects the released model (98\% and 92\% on LIBERO-Spatial) but breaks \gid (0\% on LIBERO-Spatial and LIBERO-Goal): each push's response lands on probe steps of the opposite sign, so the estimated gain can turn negative and invert the commands.
\pid finds the delay and, with the hold rule, reaches 98\% and 99\% on LIBERO-Spatial.

\begin{figure*}[t]
\centering
\includegraphics[width=\textwidth]{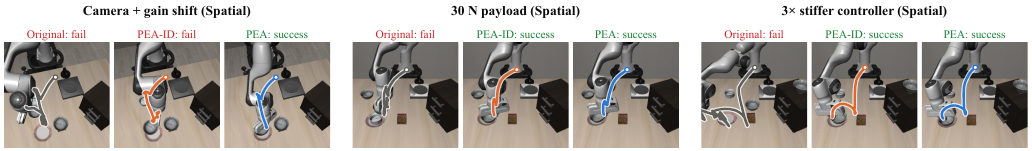}
\caption{\textbf{Rollouts from the same initial state} (Cosmos-Policy; last frame of each episode with the end-effector path, dot: start). \emph{Left}, coupled camera and gain shift: the identifier alone corrects the gains but not the view, and the adapter completes the task. \emph{Middle}, 30\,N payload: the released model presses the gripper down; the identifier's drift term lifts it. \emph{Right}, $3\times$ stiffer controller: each command moves the arm too far and the released model overshoots; the identifier reads effective gains of 1.5--1.9 and scales the commands down.}
\label{fig:rollouts}
\end{figure*}

\paragraph{Payload and stiffness.}
The drift term targets the payloads: on LIBERO-Spatial, it lifts the 15 and 30\,N configurations from 79\% and 13\% without it to 96\% and 65\%.
Stiffness enters \cref{eq:model} only approximately: the identifier fixes the stiffer controller on LIBERO-Spatial (15\% to 96\%) but lowers the softer one on LIBERO-Goal (79\% to 62\% with \pid, 55\% with \oursfull).
\Cref{fig:rollouts} illustrates a coupled shift, the 30\,N payload and the stiffer controller with one rollout each.

\begin{figure}[t]
\centering
\includegraphics[width=\linewidth]{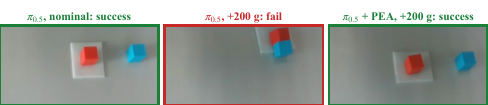}
\caption{\textbf{Real-robot rollouts on a PiPER arm} (final top-down frame of one rollout each). Under a 200\,g payload, $\pi_{0.5}$ carries the cube too low and pushes the target aside; with \ours, it places the cube as without payload.}
\label{fig:real}
\end{figure}

\subsection{Real-robot experiments}
\label{sec:res:real}
We deploy \ours on an AgileX PiPER arm with a $\pi_{0.5}$ policy~\cite{intelligence2025pi05} fine-tuned on teleoperated demonstrations of placing one cube beside another.
With the camera fixed, a 200\,g payload on the end effector acts as an invisible actuation shift; we run the full \ours, which probes the arm before each episode, compensates every command and adapts $\pi_{0.5}$ with its LoRA adapter while the base weights stay frozen.
Over 10 trials per condition, $\pi_{0.5}$ succeeds 4 times under the payload and 8 times with \ours; at this sample size, the difference is indicative rather than significant (Fisher's exact test, $p{=}0.17$).
As \cref{fig:real} shows, the payload lowers the carried cube so that it strikes the target, which \ours prevents; \cref{sec:supp:real} gives the full setup.

\subsection{Ablation and analysis}
\label{sec:res:evidence}
\label{sec:res:suites}
\label{sec:res:theory}
\label{sec:res:ablation}

\begin{figure}[!tb]
\centering
\includegraphics[width=\linewidth]{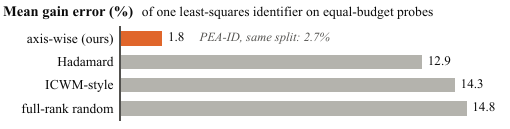}
\caption{\textbf{Probe design at equal budget.} Gain error of four 30-step probes under the diagnostic identifier of \cref{tab:supp:design}.}
\label{fig:reader}
\end{figure}

\Cref{tab:comp} adds the components of \ours to the frozen Cosmos-Policy one at a time.
We discuss LIBERO-Spatial below; LIBERO-Goal follows the same trends unless noted.

\paragraph{Each identifier term targets one kind of shift.}
As the first four rows of \cref{tab:comp} show, reading the gain alone (\gid) lifts seen types from 42.8\% to 77.3\% but lowers new dynamics from 61.0\% to 38.0\%, since the latency in these configurations biases a gain-only estimate.
Reading the delay (\pid{} w/o drift) restores new dynamics to 78.0\%, and the drift term (\pid) raises them to 89.7\%, while neither term changes seen or unseen types by more than two points.
Each term helps where its shift occurs.

\paragraph{The adapter corrects what the identifier cannot see.}
Comparing the \pid and \oursfull rows of \cref{tab:comp}, the adapter lifts seen types from 76.3\% to 93.0\% and unseen types from 68.8\% to 87.8\%.
These configurations combine actuation with camera shifts, and the identifier, which reads only motion, cannot correct the latter.
This gain does not rely on the probe: without the probe input, a LoRA of the same size (static LoRA) reaches 90.8\% and 89.3\%, on par with \oursfull within noise ($p{=}0.10$ on seen types).
Conversely, on new dynamics, which change only the actuation, the adapter adds nothing beyond \pid on LIBERO-Spatial (90.3\% vs.\ 89.7\%) and costs about four points on LIBERO-Goal.
The identifier thus corrects actuation and the adapter visual shifts, so the two are complementary.

\paragraph{The adapter reads the probe motion, not its frames.}
The last two rows of \cref{tab:comp} zero one probe input of the trained adapter at test time.
Without the probe frames, \oursfull is unchanged on all three shift types (94.5\%, 88.3\% and 89.5\% vs.\ 93.0\%, 87.8\% and 90.3\%).
Without the motion features, seen and unseen types fall to 73.5\% and 70.0\%, back to the level of \pid (76.3\% and 68.8\%): the adapter has learned to rely on these features, and removing them cancels its gain.
New dynamics stay at 90.0\%, because the identifier corrects the actuation on its own.

\paragraph{Probe design sets the identification error.}
\Cref{fig:reader} isolates the probe: four 30-step probes are read by the same least-squares identifier on 120 held-out configurations.
Our axis-wise probe yields 1.8\% mean gain error, against 12.9\%, 14.3\% and 14.8\% for Hadamard, ICWM-style and full-rank random probes, which excite all axes at once.
Exciting one axis at a time keeps each response free of the others, as \cref{prop:sep} assumes, and lets a per-axis calibration absorb the amplitude nonlinearity that \cref{eq:model} omits.
This carries over to the task: in \cref{tab:main}, swapping in the equally long ICWM-style probe lowers LIBERO-Spatial seen types from 77.3\% (\gid) to 60.5\%.

\subsection{Limitations}
\label{sec:limitations}
\ours compensates the actuation shifts that model~\eqref{eq:model} describes, namely per-axis gain, delay and drift.
Shifts outside this model, such as cross-axis coupling or loads that change within an episode, lie beyond its scope (\cref{sec:supp:mis}); detecting such a mismatch before correcting is a natural extension.
Our real-robot study covers one task and one payload, and a broader hardware evaluation is left for future work.

\section{Conclusion}
\label{sec:conclusion}
We presented \ours, a test-time calibration method for visuomotor policies under actuation shifts, which leave the observations unchanged and are therefore invisible to the policy.
Before each episode, \ours executes a 30-step probe that excites one arm axis at a time, estimates the per-axis gain, control delay and drift from the response in closed form, and compensates every subsequent command.
A LoRA adapter on the frozen policy complements this identifier for visual shifts.
On LIBERO with a WAM and a VLA, the identifier alone, without policy training, recovers most of the success lost to actuation shifts, including unseen ones, and the adapter further restores success under coupled shifts.
More broadly, when a deployment change cannot be seen, a robot should briefly interact before acting and read that interaction with an explicit model.

\section*{Acknowledgements}
This research was supported by the OpenEuroLLM project, co-funded by the Digital Europe Programme under GA no.~101195233.
Computational resources for this work were provided by the Spike-1 GPU cluster, operated by the Supercomputing Center at Eindhoven University of Technology.

\section*{LLM Usage Statement}
Generative AI tools were used for language polishing, \mbox{LaTeX} formatting and code development, but not for research ideation or data generation.
The authors reviewed the edits and take responsibility for the final manuscript.

{
    \small
    \bibliographystyle{ieeenat_fullname}
    \bibliography{main}
}

\clearpage
\appendix
\raggedbottom
\setcounter{figure}{0}\setcounter{table}{0}
\renewcommand{\thefigure}{S\arabic{figure}}
\renewcommand{\thetable}{S\arabic{table}}
\renewcommand{\theHfigure}{S\arabic{figure}}
\renewcommand{\theHtable}{S\arabic{table}}
\renewcommand{\topfraction}{0.9}\renewcommand{\bottomfraction}{0.6}\renewcommand{\textfraction}{0.1}
\renewcommand{\floatpagefraction}{0.75}\renewcommand{\dbltopfraction}{0.9}\renewcommand{\dblfloatpagefraction}{0.75}
\setcounter{topnumber}{3}\setcounter{dbltopnumber}{3}\setcounter{totalnumber}{5}
\makeatletter\setlength{\@fptop}{0pt}\setlength{\@fpsep}{14pt}\setlength{\@fpbot}{0pt plus 1fil}\setlength{\@dblfptop}{0pt}\setlength{\@dblfpsep}{14pt}\setlength{\@dblfpbot}{0pt plus 1fil}\makeatother
\renewcommand{\contentsname}{Appendix}
\addtocontents{toc}{\protect\setcounter{tocdepth}{2}}

{\small
\tableofcontents
}
\vspace{0.5em}
\vspace{0.5em}
\noindent This appendix supplements the main paper.
Figures and tables numbered S1, S2, \dots\ belong to this appendix; all other numbers refer to the main paper.
\Cref{sec:supp:setup,sec:supp:impl,sec:supp:baselines} give the setup, the implementation and the baselines; \cref{sec:supp:ident,sec:supp:return} analyze the probe and the identifier; \cref{sec:supp:lora} motivates the LoRA adapter; \cref{sec:supp:results,sec:supp:real} report further results, and \cref{sec:supp:qual,sec:supp:cells} give rendered examples and per-configuration results.

\section{Experimental setup}
\label{sec:supp:setup}
This section details the setup summarized in \cref{sec:setup}.

\paragraph{Training shifts.}
Each training episode is nominal with probability 0.1 and otherwise draws independent per-axis action gains $g_i\sim\mathcal{U}[0.7,2.0]$, a camera rotation and a gripper force limit of $\mathcal{U}[0.25,1.0]$ times nominal.
The agent-view camera rotates about the point at robot-base height on its optical axis (azimuth ${\sim}\,\mathcal{U}[-40^\circ,40^\circ]$, elevation ${\sim}\,\mathcal{U}[0^\circ,25^\circ]$).

\paragraph{Test configurations.}
\label{sec:supp:cfg}
Each test configuration is fixed for all of its episodes; \cref{tab:supp:cfg} lists them.
Seen types (s1--s6) are new draws of the training shift types.
Unseen types (u1--u4) add types absent from training: a $1.5\times$ zoom-out (u1), object masses $\times100$ (u2) or both (u3) on a seen-type coupling, or only zoom-out, mass $\times100$ and a field of view $\times0.8$ (u4).
New dynamics (d1--d6) each change one actuation factor that training never varies.
OSC stiffness $k_p$ is scaled by 0.3 (d1) or 3 (d2), with the damping scaled by the square root of the factor.
An uncompensated constant downward force of 15\,N (d3) or 30\,N (d4) pulls the end effector.
Latency delays every command by 1 (d5) or 2 (d6) control steps (50 or 100\,ms).
\Cref{sec:supp:results} further reports combined shifts (c1--c4), extrapolation beyond the training range (x1--x4), shifts outside model~\eqref{eq:model} (m1--m4) and four LIBERO-RIG~\cite{cui2026liberorig} axes.

\begin{table}[t]
\centering
\scriptsize
\setlength{\tabcolsep}{2.2pt}
\resizebox{\linewidth}{!}{%
\begin{tabular}{l cccccc cc c l}
\toprule
 & $g_x$ & $g_y$ & $g_z$ & $g_{rx}$ & $g_{ry}$ & $g_{rz}$ & az & el & grip & unseen types \\
\midrule
s1 & 1.71 & 1.67 & 1.76 & 0.98 & 0.78 & 1.82 & $-0.4$ & 3.7 & 0.49 & -- \\
s2 & 0.97 & 1.02 & 0.82 & 0.99 & 1.57 & 0.98 & $-36.0$ & 4.0 & 0.98 & -- \\
s3 & 1.67 & 1.05 & 1.23 & 2.00 & 1.97 & 0.97 & 14.9 & 23.0 & 0.80 & -- \\
s4 & 1.99 & 1.74 & 1.73 & 0.82 & 1.46 & 0.76 & 39.9 & 24.9 & 0.41 & -- \\
s5 & 1.94 & 0.92 & 1.44 & 1.93 & 0.82 & 0.76 & $-32.4$ & 4.8 & 0.32 & -- \\
s6 & 1.60 & 0.95 & 1.42 & 1.22 & 1.40 & 1.36 & 9.1 & 16.8 & 0.55 & -- \\
\midrule
u1 & 0.97 & 1.55 & 1.07 & 1.32 & 1.29 & 1.30 & $-4.4$ & 5.4 & 0.89 & zoom 1.5 \\
u2 & 1.38 & 0.78 & 1.71 & 1.59 & 1.41 & 1.83 & $-26.2$ & 2.9 & 0.94 & mass 100 \\
u3 & 1.80 & 1.52 & 1.30 & 1.71 & 1.34 & 0.91 & $-39.1$ & 20.4 & 0.76 & zoom, mass \\
u4 & 1 & 1 & 1 & 1 & 1 & 1 & 0 & 0 & 1 & zoom, mass, FoV 0.8 \\
\bottomrule
\end{tabular}}
\caption{\textbf{Test couplings.} Seen types s1--s6 and unseen types u1--u4: per-axis action gains, camera azimuth and elevation in degrees, and gripper force as a fraction of nominal. The new dynamics d1--d6 keep nominal gains, camera and gripper force and change one actuation factor each.}
\label{tab:supp:cfg}
\end{table}

\paragraph{Policies and data.}
We use the released Cosmos-Policy-LIBERO-Predict2-2B checkpoint~\cite{kim2026cosmospolicy} and the released four-suite OpenVLA-OFT checkpoint~\cite{kim2025oft}, which reproduces 98\% and 97\% nominal success on LIBERO-Spatial and LIBERO-Goal in our evaluator.
Adapters and fine-tuned bases are trained once per suite pair, LIBERO-Spatial with LIBERO-Goal and LIBERO-Object with LIBERO-Long.
Training data are the shifted images, probes and actions of successful \teach episodes from initial states 10--49, disjoint from the test states (697 and 746 of 800 episodes for the two suite pairs).

\paragraph{Compared systems.}
An \emph{estimator} corrects commands from interaction data without learned weights.
The first group of \cref{tab:main} keeps the released weights frozen.
\raw is the released model; \pid adds our identifier, and \pid{} w/o drift omits the drift term of \cref{eq:bias}.
\gid divides each axis's mean signed probe response by its nominal value and reads neither delay nor drift.
ICWM-style~\cite{wang2026icwm} reads the gains from a random 30-step probe, and SPACE-style~\cite{lee2026space} fits an affine command adapter to 500 random calibration steps; ``-style'' marks our re-implementation of their evidence source, not the published systems.
\oursfull adds the LoRA adapter to \pid, and \pid{} + prefix adapter writes 16 probe tokens into the instruction embedding instead.
The second group fine-tunes the base on intended labels without probe input (FT base) and runs the estimators on it; FT + ARX identifies an ARX model from the 500-step calibration (\cref{sec:supp:sysid}), and \oursfull-FT trains the adapter jointly with the base.
\Cref{sec:supp:lora,tab:abl} also compare learned readers without an estimator.

\paragraph{Training.}
All trained Cosmos-Policy models use the same data, 6,000 steps and a global batch of 30.
New modules train at a learning rate of $10^{-4}$, and fine-tuned bases at $6\times10^{-6}$, the final-stage rate of the released recipe.

\paragraph{Statistics.}
Success rates (in \%) average 100 episodes per configuration and suite: 10 tasks $\times$ initial states 0--9.
Per configuration, the 95\% Wilson interval~\cite{wilson1927probable} spans at most $\pm10$ points, and the standard error of a six-configuration mean is about 2 points.
Exact McNemar tests pair episodes by configuration, task and initial state.

\section{Implementation details}
\label{sec:supp:impl}

\paragraph{Adapter.}
A probe encoder (26.4M parameters) maps the probe to a code $z$.
A numeric branch (LayerNorm, linear layer, GELU) embeds the per-step 15-d feature $[a_t,\,100\Delta p_t,\,10\,\mathrm{rotvec}(\Delta R_t),\,100\Delta q_t]$ ($q$: finger positions) of the 30 probe steps.
An image branch stacks the first of 31 agent-view frames ($64{\times}64$) with the 30 differences to it and applies four stride-2 convolutions, so it sees how the known probe motion appears from the current camera; an MLP fuses the two 1024-d features.
The adapted layers of \cref{eq:lora} are the self-attention $q,k,v$ and output projections and both MLP layers of the 28 DiT blocks (168 layers).
A zero-initialized linear head generates the spectrum $s(z)\in\mathbb{R}^{168\times14}$, so training starts exactly at the released model; the factors $A_l,B_l$ learn directions shared by all episodes, and each episode's adaptation is 2,352 numbers.
The adapter has 43.3M parameters, 16.9M besides the encoder.
After training, all 715 DiT tensors are bit-identical to the released checkpoint.

\paragraph{Labels.}
The \teach executes the compensated arm action $\bar a/g$: the released model's arm action $\bar a$, computed from the nominal camera view, divided by the true per-axis gain $g$.
\oursfull, \oursfull-FT and the FT base train on the intended label $\mathrm{clip}\big((\bar a/g)\odot g,\,-1,\,1\big)$, the \teach's arm action in nominal space; the learned readers without an estimator train on the compensated label $\bar a/g$, with the same gripper label.

\paragraph{Probe protocol.}
The probe runs after the environment is reset to the episode's initial state and before the first policy step, with the episode's shift active, including latency and payload.
In the main evaluation, we snapshot the simulator and controller states before the probe and restore both after it, so the episode starts from exactly the pre-probe state; \cref{sec:supp:return} replaces this reset with a physical return.
All probe-based methods use the same protocol.

\paragraph{Identifier.}
The templates $T_i$, the calibration $\phi_i$ and the drift statistics $\mu_i,\sigma_i$ are computed once from the 800 training probes.
The nominal motion per unit command, $k_i$, is the diagonal of an affine fit of per-step displacement on command over nominal episodes.
Nothing is fitted on test configurations or per policy: Cosmos-Policy and OpenVLA-OFT use the same identifier.
Training contains no latency, so the test latencies of 1 and 2 steps lie outside the data the identifier was built from.

\paragraph{Baseline estimators.}
\gid divides each axis's mean signed probe response $r_i$ by its mean $\bar r_i$ over nominal training probes and maps the ratio to a gain with a per-axis quadratic calibration fitted like $\phi_i$.
The ICWM-style probe keeps $K{=}30$ steps but consists of five 6-step segments, each along a random unit direction in the 6-D arm action space scaled by 0.5, with the gripper open; the direction seed depends on suite, task and initial state.
Its gain per axis is the least-squares slope of the axis's motion on its command, dropping the first two steps of each segment, followed by the same calibration; this reading was the best of the variants we tried on the training probes.
The SPACE-style baseline~\cite{lee2026space} runs 10 calibration trajectories of 50 steps from the initial state, re-mixing a random direction (70\% old, 30\% new) every 10 steps, and fits an affine adapter $u=W\Delta p+b$ by least squares on the 500 (displacement, command) pairs, updated online by least-mean-squares with step size 0.2.
Because our policies output commands rather than displacements, the intended displacement is the nominal response $\Delta p=Aa+c$ of the command $a$, with $A$ and $c$ fitted once on nominal episodes.

\paragraph{Learned readers.}
The learned readers of \cref{tab:abl,tab:reader} fine-tune the base on compensated labels and condition it on evidence through the probe encoder above.
Prefix tokens overwrite the last 16 positions of the T5 instruction embedding, which are zero padding for every LIBERO instruction; FiLM modulates the normalized inputs of all attention and MLP layers; probe-LoRA uses the adapter of \cref{eq:lora}.
Each head adds 16.8--16.9M parameters and is zero-initialized.
The history reader reads the current episode's last 30 steps and 16 frames (Reflective-VLA-style), the earlier-episode reader a released-model episode under the same shift from another initial state (Self-Adaptive-VLA-style), and the random-probe reader the ICWM-style probe.
\learn reads the scripted probe through prefix tokens, with the encoder trained at the base learning rate.

\paragraph{OpenVLA-OFT.}
We use the released four-suite checkpoint~\cite{kim2025oft} with the upstream LIBERO evaluation settings.
The identifier and the hold rule are applied exactly as for Cosmos-Policy.
Its weights are never updated; the adapter adapts the 224 linear layers of its Llama backbone and is trained per suite pair.

\section{Baseline details}
\label{sec:supp:baselines}

\subsection{SPACE-style adapter}
\label{sec:supp:space}
To see why the SPACE-style rows lose nominal success, we composed the fitted adapter $W$ with the nominal command-to-displacement map $A$; under nominal dynamics, the ideal product is the identity.
For the inverse regression of SPACE (command on displacement), the diagonal of $WA$ is 0.75--0.90: displacements lag behind the commands, which attenuates the regression slope, so the adapter under-commands.
Off-diagonal terms reach 0.5 and mix axes that the controller keeps separate.
Fitting the forward map (displacement on command) and inverting it brings the diagonal to 0.95--1.09, but the cross terms remain.
SPACE retrains its policy to output displacements, whereas our policies output commands, so these numbers characterize the adapter in our command-space setting, not SPACE as published.

\subsection{System-identification baseline (FT + ARX)}
\label{sec:supp:sysid}
FT + ARX is a conventional system-identification baseline on the FT base.
\textbf{Model.} For every end-effector displacement channel $i$ and all six arm commands $j$,
$y^i_t = a^i_1 y^i_{t-1} + a^i_2 y^i_{t-2} + \sum_j (b^{ij}_0 u^j_{t-n_k} + b^{ij}_1 u^j_{t-n_k-1}) + c^i + e^i_t$, with a common input delay $n_k\in\{0,\dots,3\}$.
\textbf{Identification.} Per channel, the parameters get a Gaussian prior centered on the nominal model; the prior variances, the noise variance and $n_k$ maximize the marginal likelihood~\cite{ljung1999system,pillonetto2014kernel}.
\textbf{Compensation.} With static gain $G$ and static offset $c_{ss}$, the arm command becomes $u' = G^{-1}G_{dc}\,u + b$, with singular values clipped to $[0.25, 4]$ and $b = -G^{-1}(c_{ss}-c_0)$ on channels that deviate by more than three nominal standard deviations; after each chunk, the arm holds for $n_k - n_{k,0}$ steps.
\textbf{Online refinement.} Recursive least squares with forgetting factor 0.995 refines the compensation during the episode.
\textbf{Excitation.} FT + ARX in \cref{tab:main} uses the 500-step calibration; \cref{tab:abl} also identifies it from our 30-step probe.

The full $6\times6$ model makes the estimate less precise than our per-axis identifier.
On 200 training probes, its per-axis gain error after compensation is 20.5, 16.4, 8.4, 9.6, 76.7 and 32.6\% ($x,y,z,r_x,r_y,r_z$), against 3.7, 0.3, 0.8, 0.8, 9.1 and 1.5\% for our identifier, while both recover the delay on every probe.
Under nominal dynamics, the online refinement also drifts away from the identity, most in failed episodes (\cref{tab:supp:sysidmech}).

\begin{table}[t]
\centering
\small
\setlength{\tabcolsep}{4pt}
\begin{tabular}{l l c c c c}
\toprule
Excitation & Episodes & $n$ & diag. & off-diag. & bias \\
\midrule
30-step probe & Spatial, success & 90 & 0.24 & 0.85 & 32\% \\
 & Spatial, failure & 10 & 0.25 & 1.08 & 57\% \\
 & Goal, success & 69 & 0.39 & 1.15 & 24\% \\
 & Goal, failure & 31 & 1.52 & 2.32 & 74\% \\
\midrule
500-step calib. & Spatial, success & 95 & 0.11 & 0.36 & 0\% \\
 & Spatial, failure & 5 & 0.17 & 0.67 & 1\% \\
 & Goal, success & 83 & 0.23 & 0.62 & 3\% \\
 & Goal, failure & 17 & 1.12 & 1.84 & 39\% \\
\bottomrule
\end{tabular}
\caption{\textbf{Online drift of FT + ARX under nominal dynamics}, where the ideal compensation $M$ is the identity and needs no bias. diag.: $\max_i|M_{ii}{-}1|$ and off-diag.: largest off-diagonal entry, both at the end of the episode; bias: share of steps with a bias.}
\label{tab:supp:sysidmech}
\end{table}

\section{Identification analysis}
\label{sec:supp:ident}
This section analyzes how accurately the probe identifies the actuation, why moving one axis at a time helps, and how the identifier behaves on probes it was not fitted on.

\begin{figure*}[t]
\centering
\includegraphics[width=\textwidth]{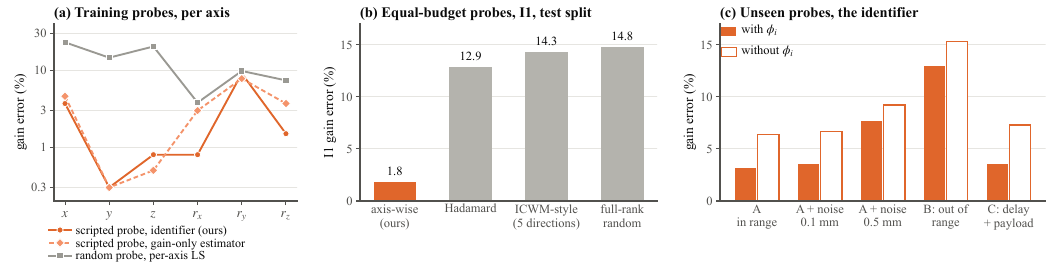}
\caption{\textbf{Identification accuracy.} (a)~Gain error per axis on the training probes for three estimators. (b)~Four equal-budget probes read by one diagnostic least-squares identifier (\cref{tab:supp:design}). (c)~Our identifier on unseen probes with and without the calibration $\phi_i$ (\cref{tab:supp:calcheck}).}
\label{fig:supp:ident}
\end{figure*}

\subsection{Identification accuracy}
\label{sec:supp:est}
\Cref{fig:supp:ident} summarizes the identification accuracy.
On the 800 training probes, which contain no latency, the identifier selects $\hat d{=}0$ for every probe.
The ICWM-style random probe, in contrast, gives noisy gain readings even under nominal gains: its least-squares slope varies with a coefficient of variation of 58\% ($x$), 23\% ($y$) and 43\% ($z$).

\subsection{Probe design at equal budget}
\label{sec:supp:design}
\Cref{tab:supp:design} gives the numbers behind \cref{fig:reader}.
\textbf{Designs.} All probes have 30 steps with arm commands of norm 0.5.
Ours pushes one axis at a time ($+,+,-,-$).
The ICWM-style probe holds random directions in rank-5 segments.
The full-rank random probe applies 12 random directions and then their negations in reverse order, and the Hadamard probe does the same with the sign patterns of six orthogonal columns of a Hadamard matrix.
\textbf{Identifiers.} The diagnostic identifier I1 fits model~\eqref{eq:model} to any excitation, $y_t=\sum_j g_j Z_j(t-d)+c$, where $Z_j$ is the nominal response to axis $j$'s commands alone, predicted by the ARX model of \cref{sec:supp:sysid} identified on nominal data; it solves least squares for each $d\in\{0,\dots,3\}$ and keeps the delay with the smallest residual.
I2 is the regularized ARX identification of \cref{sec:supp:sysid}.
Each has a per-axis quadratic calibration.
\textbf{Data.} 240 configurations (per-axis gains log-uniform in $[0.7,2]$, delays of 0--2 steps, random initial states), the same for every design; 120 calibrate and 120 test.

\begin{table}[t]
\centering
\small
\setlength{\tabcolsep}{4pt}
\begin{tabular}{l c c c}
\toprule
Probe & I1 gain error & I2 gain error & delay found \\
\midrule
axis-wise (ours) & \textbf{1.8\%} & \textbf{10.9\%} & 100\% \\
Hadamard & 12.9\% & 16.0\% & 100\% \\
ICWM-style & 14.3\% & 18.5\% & 97\% \\
full-rank random & 14.8\% & 18.3\% & 100\% \\
\bottomrule
\end{tabular}
\caption{\textbf{Identification error of four equal-budget probes} (120 test configurations; mean absolute relative error over the six gains; delay found by I1). On our probe, the identifier of \cref{sec:method:id} has 2.7\% gain error on the same split.}
\label{tab:supp:design}
\end{table}

\subsection{Superposition and amplitude nonlinearity}
\label{sec:supp:superpos}
Moving one axis at a time wins because a per-axis calibration can then absorb the amplitude nonlinearity of the response.
For 40 configurations (gains log-uniform in $[0.7,2]$), we ran each probe, each of its six single-axis components alone and the probe with all arm commands set to zero, all from the same state.
Regressing the response on the measured component responses, which removes any error of the nominal model, gives per-axis gain errors of 9.7\% for the Hadamard probe and 11.5\% for ours before calibration; our errors concentrate on $r_x$ and $r_y$ (25 and 37\%, translations at most 1\%).
The controller responds nonlinearly to the amplitude: doubling the command of our probe from 0.5 to 1.0 scales the translation responses by 1.97--2.12 but the rotation responses by only 1.27--1.70.
With one axis at a time, each axis's nonlinearity depends only on its own command, so the per-axis calibration absorbs it (1.8\% after calibration, \cref{tab:supp:design}); with all axes moving, it depends on the other axes' commands (12.9\% for the Hadamard probe).

\subsection{Calibration on unseen probes}
\label{sec:supp:calcheck}
\Cref{tab:supp:calcheck} applies the identifier, unchanged, to 360 new probes (120 per set, random initial states).
Set A has in-range gains (log-uniform in $[0.7,2]$) with delays of 0--2 steps, also with Gaussian displacement noise (0.1 and 0.5\,mm per step; a unit command moves the end effector about 12\,mm per step).
Set B has gains outside the training range (each axis in $[0.4,0.6]$ or $[2.2,3]$), and set C combines in-range gains with delays and a payload of 0, 15 or 30\,N.
The delay is found on every probe, and the per-axis calibration halves the gain error in range.
The 3.2\% of set A differs from the 2.7\% of \cref{tab:supp:design} because the two use different probe sets and calibration data.

\begin{table}[t]
\centering
\small
\setlength{\tabcolsep}{4pt}
\begin{tabular}{l c c c}
\toprule
Probe set & with $\phi_i$ & without $\phi_i$ & delay found \\
\midrule
A in range & \textbf{3.2\%} & 6.4\% & 100\% \\
A + 0.1\,mm noise & \textbf{3.6\%} & 6.7\% & 100\% \\
A + 0.5\,mm noise & \textbf{7.7\%} & 9.2\% & 100\% \\
B out of range & \textbf{13.0\%} & 15.3\% & 100\% \\
C delay + payload & \textbf{3.6\%} & 7.3\% & 100\% \\
\bottomrule
\end{tabular}
\caption{\textbf{Gain error of the identifier on unseen probes} (mean absolute relative error over the six axes, 120 probes per set), with and without the per-axis calibration $\phi_i$ fitted on the 800 training probes. In set C, the error is 4.4\% under the 30\,N payload.}
\label{tab:supp:calcheck}
\end{table}

\section{Physical return after the probe}
\label{sec:supp:return}
Our main evaluation restores the pre-probe simulator state; here we measure what a physical return costs instead.
\textbf{Controller.} With pose error $e$ (position, rotation vector) to the pre-probe pose, the arm command is $\operatorname{clip}\big(\tfrac{0.5}{1+\hat d}\,e/k/\hat g + \hat b,\,-1,\,1\big)$, with $k$ the nominal per-step response to a unit command (0.0122\,m, 0.12\,rad) and $\hat g,\hat d,\hat b$ from the probe.
The gripper stays open, and the loop stops within 2\,mm and 0.02\,rad or after 40 steps; the gain is reduced with the delay because holding after each command would overshoot through the controller's own lag.
\textbf{Cost.} \Cref{tab:supp:ret} reports three tasks, six conditions and five initial states each.
The return takes at most 25 steps, ends within 2\,mm of the start and moves no object; object displacement is measured against an idle rollout of equal length from the same state.

\begin{table}[t]
\centering
\footnotesize
\setlength{\tabcolsep}{3pt}
\begin{tabular}{l c c c c}
\toprule
Condition & excursion & return steps & residual & objects moved \\
\midrule
nominal & 19--20 & 6.0--7.6 & 1.3--1.6 & 0 \\
per-axis gains & 22--31 & 5.7--7.8 & 1.2--1.8 & 0 \\
stiffness $\times0.3$ & 9--11 & 5.3--6.6 & 0.8--1.6 & 0 \\
stiffness $\times3$ & 35--36 & 3.3--4.0 & 1.1--1.3 & 0 \\
payload 30\,N & 113--115 & 17.8--19.4 & 1.4--1.6 & 0 \\
latency 2 steps & 19--20 & 13.0--15.0 & 1.6--1.8 & 0 \\
\bottomrule
\end{tabular}
\caption{\textbf{Cost of the physical return} (excursion, residual and object motion in mm). Ranges over three tasks (LIBERO-Spatial task 0, LIBERO-Goal tasks 2 and 5) with five initial states each; return steps: mean per task, at most 25 in any episode; excursion: largest end-effector distance from the start during the probe.}
\label{tab:supp:ret}
\end{table}

\section{Why a LoRA adapter: the gain-compensation test}
\label{sec:supp:lora}
This section compares the interfaces through which the probe can enter the policy and thereby motivates the LoRA adapter of \cref{sec:method:lora}.

\paragraph{Representational argument.}
A LoRA with a probe-generated spectrum contains gain compensation as an exact special case, whereas prefix tokens and FiLM do not.
Compensating gains $g$ means mapping every arm action $a=Wh$ of a linear read-out to $\mathrm{diag}(1/g)\,a$, whatever the hidden state $h$.
In the family of \cref{eq:lora}, a rank-6 update does this exactly: with $A$ the six arm rows of $W$, $B$ the matching selection and $s_i(z)=1/g_i-1$, $W(z)h=\mathrm{diag}(1/g)\,Wh$ for all $h$, and each spectrum entry is a scalar function of one axis's gain, which the probe determines.
Prefix tokens give an attention output $(1-\lambda)\,o_c+\lambda\,o_p$, with $o_c$ the output over the content, $o_p$ a convex combination of the prefix values and $\lambda$ the attention mass on the prefix~\cite{petrov2024prompting}, so the residual stream moves along directions fixed by $z$.
FiLM~\cite{perez2018film}, $h\mapsto\mathrm{diag}(1+\gamma(z))\,h+\beta(z)$, scales fixed feature coordinates, and $W\mathrm{diag}(1+\gamma)=\mathrm{diag}(1/g)\,W$ has no solution unless the arm read-out aligns with them.
Both must approximate the scaling through later nonlinear layers.
Cosmos-Policy decodes its action from denoised latents, so the construction is not exact there; the argument concerns what each interface can represent.

\paragraph{Test.}
To test this, we remove the estimator: the base is fine-tuned on compensated labels, so the policy must compensate the gain itself.
The four learned interfaces share the probe encoder, data, schedule and parameter budget; DR-only reads no evidence.
We measure compensation as the slope of the log mean action magnitude against the log true gain over the couplings s1--s6 and u1--u3; a slope of $-1$ means full compensation.

\begin{figure}[t]
\centering
\includegraphics[width=\linewidth]{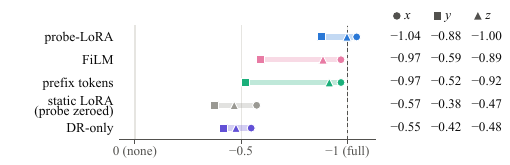}
\caption{\textbf{Gain compensation without an estimator.} Slope of action magnitude against true gain on log scales, per translation axis; $-1$ means full compensation.}
\label{fig:supp:comp}
\end{figure}

\begin{table*}[t]
\centering
\small
\setlength{\tabcolsep}{6pt}
\begin{tabular}{l c c c c}
\toprule
System & Seen types & Unseen types & New dynamics & Slope $x/y/z$ \\
\midrule
\multicolumn{5}{l}{\textit{Gain-compensation test: base fine-tuned on compensated labels, no estimator}} \\
DR-only & \sg{46.2}{52.3} & \sg{55.8}{51.8} & \sg{49.5}{62.2} & $-0.55/{-0.42}/{-0.48}$ \\
static LoRA (probe zeroed) & \sg{45.5}{49.3} & \sg{53.5}{53.2} & \sg{50.3}{62.3} & $-0.57/{-0.38}/{-0.47}$ \\
prefix tokens & \sg{85.0}{77.7} & \sg{75.3}{72.8} & \sg{51.8}{61.2} & $-0.97/{-0.52}/{-0.92}$ \\
FiLM & \sg{85.8}{80.7} & \sg{79.5}{68.5} & \sg{53.5}{61.5} & $-0.97/{-0.59}/{-0.89}$ \\
probe-LoRA & \sg{\textbf{92.5}}{\textbf{87.2}} & \sg{\textbf{89.5}}{\textbf{82.8}} & \sg{\textbf{54.5}}{58.2} & $\mathbf{-1.04}/\mathbf{-0.88}/\mathbf{-1.00}$ \\
\midrule
\multicolumn{5}{l}{\textit{Full system: base frozen, identifier in front, intended labels}} \\
\pid{} (no adapter) & \sg{76.3}{82.3} & \sg{68.8}{68.3} & \sg{89.7}{\textbf{84.0}} & -- \\
\pid{} + prefix adapter & \sg{63.0}{39.5} & \sg{63.5}{50.0} & \sg{74.2}{56.0} & -- \\
\oursfull{} (LoRA adapter) & \sg{\textbf{93.0}}{\textbf{91.5}} & \sg{\textbf{87.8}}{\textbf{80.0}} & \sg{\textbf{90.3}}{79.7} & -- \\
\bottomrule
\end{tabular}
\caption{\textbf{How the probe enters the policy} (Cosmos-Policy; success \%, LIBERO-Spatial\,/\,LIBERO-Goal). Upper block: without an estimator, the policy must compensate the gain itself; the learned interfaces share the encoder, data, schedule and parameter budget (16.8--16.9M). Lower block: the full system, where the identifier compensates the actuation. Bold: best per block and column.}
\label{tab:reader}
\end{table*}

\begin{figure}[t]
\centering
\includegraphics[width=\linewidth]{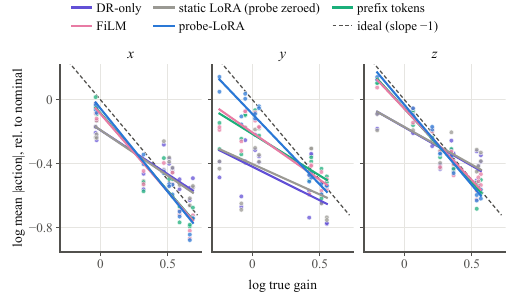}
\caption{\textbf{Gain compensation per coupling} without an estimator. Each point is one coupling of one suite (s1--s6, u1--u3; LIBERO-Spatial and LIBERO-Goal): log true gain against log mean $|$action$|$ relative to nominal. Lines: episode-level slopes of \cref{tab:reader} through each system's centroid; dashed: full compensation (slope $-1$).}
\label{fig:supp:slopes}
\end{figure}

\paragraph{Results.}
\Cref{tab:reader} and \cref{fig:supp:comp} confirm the argument.
Probe-LoRA compensates the gain almost fully (slopes $-1.04$, $-0.88$, $-1.00$), prefix tokens and FiLM only partly, and the static LoRA, the same module with its probe input zeroed, no more than DR-only.
Probe-LoRA is also the most successful interface on seen and unseen types; paired over the 3,400 episodes of all configurations, it beats prefix tokens 380 to 177 episodes ($p{<}10^{-3}$) and FiLM 334 to 170 ($p{=}2\times10^{-13}$).
The same holds on LIBERO-Object and LIBERO-Long (slopes $-0.93$, $-0.99$, $-1.01$; 260 vs.\ 180 wins over prefix tokens, $p{=}1.6\times10^{-4}$).
\Cref{fig:supp:slopes} shows the compensation per coupling.

Without an estimator, however, no interface handles the 30\,N payload or the 2-step latency, which training never shows, and all stay below the released model on new dynamics (\cref{tab:reader}; released model: 61.0\% and 69.7\%).
This is why \ours places the explicit identifier in front of the policy and trains the adapter on intended labels: with the identifier, the LoRA adapter adds the visual robustness while keeping the new-dynamics success of \pid on LIBERO-Spatial, whereas a prefix adapter lowers every column.

\section{Additional results}
\label{sec:supp:results}

\subsection{New dynamics}
\label{sec:supp:dynres}
\begin{figure*}[t]
\centering
\includegraphics[width=\textwidth]{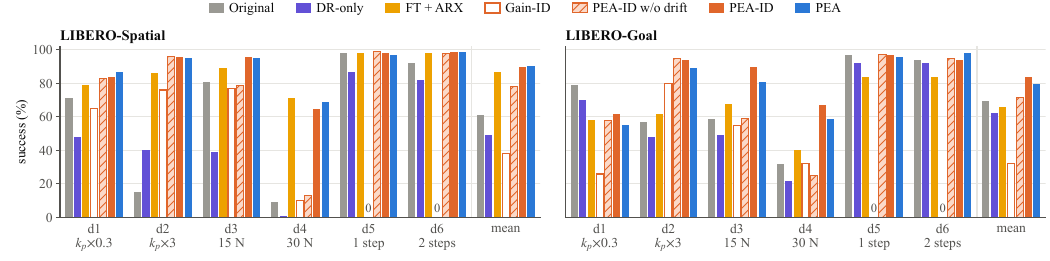}
\caption{\textbf{New dynamics, absent from training} (Cosmos-Policy, nominal camera and gains, $n{=}100$ per bar). Reading only the gain (\gid) scores 0 under latency (d5, d6); reading the delay restores these configurations, and the drift term lifts the payloads (d3, d4).}
\label{fig:dyn}
\end{figure*}

\begin{figure*}[t]
\centering
\includegraphics[width=\textwidth]{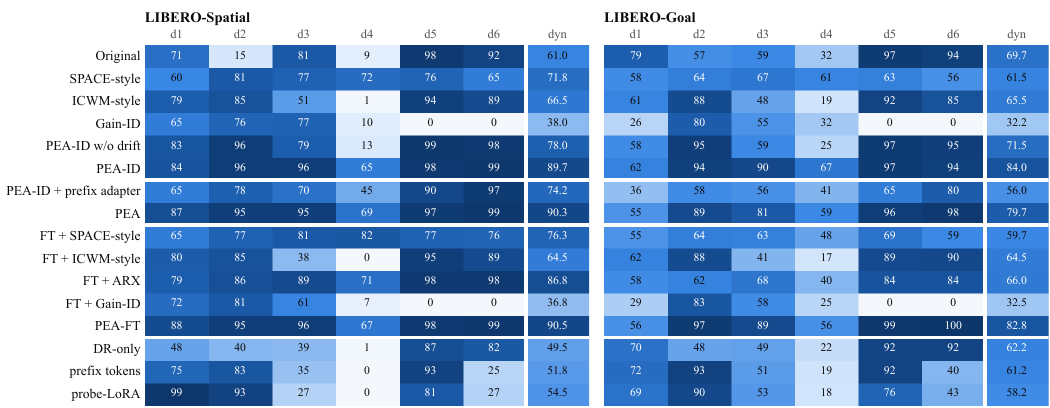}
\caption{\textbf{Per-configuration success under the six new dynamics} (Cosmos-Policy, \%, $n{=}100$ per cell; dyn: mean over d1--d6). d1, d2: stiffness $k_p\times0.3$, $\times3$; d3, d4: payload 15, 30\,N; d5, d6: latency 1, 2 steps. Groups from top: released weights with a training-free estimator; frozen base with an adapter; FT base with an estimator, and \oursfull-FT; base fine-tuned on compensated labels, no estimator.}
\label{fig:supp:dyn}
\end{figure*}

\Cref{fig:dyn} shows the six new dynamics for the released model and the identifier variants, and \cref{fig:supp:dyn} gives every system of \cref{tab:main} per configuration.
Reading only the gain (\gid) fails under latency, because each push's response lands on probe steps of the opposite sign; reading the delay restores these configurations, and the drift term lifts the payloads.
Among the frozen baselines, the ICWM-style random probe fails under the 30\,N payload (1\% and 19\%), which its gain reading does not account for.

\subsection{Combined shifts, extrapolation and model mismatch}
\label{sec:supp:comb}
\label{sec:supp:mis}
\Cref{tab:supp:comb} adds latency, a payload or both to seen-type couplings.
The identifier and the adapter compose: \pid handles the actuation part, and the adapter lifts the configurations with the largest camera changes (c2 and c4 build on s4 and s3).
\Cref{tab:supp:extra} pushes the seen types beyond the training range (x1--x4); since the identifier reads the gains from the probe rather than from training data, \oursfull extends well beyond the training gains.
Four further shifts violate model~\eqref{eq:model}: m1 rotates the $x/y$ translation commands by 30$^\circ$ (cross-axis mixing), m2 executes half of each command one step and half two steps late, m3 removes 0.15 from every arm command's magnitude (dead zone), and m4 ramps a downward force from 0 to 30\,N during the task.
The identifier reads each as the nearest shift of its model, \eg, m3 as gains of about 0.7.
This helps under the dead zone, is neutral under the fractional delay and the force ramp, and hurts under cross-axis mixing, which no per-axis model captures.

\begin{table}[t]
\centering
\small
\setlength{\tabcolsep}{4pt}
\begin{tabular}{l ccc c}
\toprule
 & \raw & \pid & \oursfull & \oursfull-FT \\
\midrule
c1 & \sg{14}{34} & \sg{96}{\textbf{99}} & \sg{\textbf{98}}{97} & \sg{99}{99} \\
c2 & \sg{31}{13} & \sg{52}{59} & \sg{\textbf{86}}{\textbf{80}} & \sg{87}{80} \\
c3 & \sg{20}{26} & \sg{\textbf{93}}{\textbf{93}} & \sg{90}{84} & \sg{89}{85} \\
c4 & \sg{49}{33} & \sg{36}{61} & \sg{\textbf{76}}{\textbf{83}} & \sg{83}{91} \\
\midrule
mean & \sg{28.5}{26.5} & \sg{69.3}{78.0} & \sg{\textbf{87.5}}{\textbf{86.0}} & \sg{89.5}{88.8} \\
\bottomrule
\end{tabular}
\caption{\textbf{Combined shifts} (Cosmos-Policy; success \%, LIBERO-Spatial\,/\,LIBERO-Goal). c1: s1 + latency 1; c2: s4 + 15\,N; c3: s1 + latency 2 + 15\,N; c4: s3 + latency 1 + 15\,N. Bold: best with the released weights frozen.}
\label{tab:supp:comb}
\end{table}

\begin{table}[t]
\centering
\small
\setlength{\tabcolsep}{4pt}
\begin{tabular}{l l ccc}
\toprule
 & Shift & \raw & \oursfull & \oursfull-FT \\
\midrule
x1 & gains 2.1--2.6 & \sg{15}{33} & \sg{92}{89} & \sg{91}{94} \\
x2 & gains 0.4--0.6 & \sg{20}{38} & \sg{69}{64} & \sg{78}{60} \\
x3 & camera & \sg{60}{42} & \sg{96}{76} & \sg{96}{93} \\
x4 & gains + camera & \sg{1}{1} & \sg{33}{47} & \sg{40}{51} \\
\midrule
m1 & cross-axis mixing & \sg{27}{29} & \sg{14}{15} & \sg{10}{26} \\
m2 & 1.5-step delay & \sg{91}{94} & \sg{88}{87} & \sg{88}{93} \\
m3 & dead zone & \sg{59}{63} & \sg{79}{65} & \sg{86}{72} \\
m4 & force ramp & \sg{96}{68} & \sg{96}{71} & \sg{95}{68} \\
\bottomrule
\end{tabular}
\caption{\textbf{Extrapolation and shifts outside model~\protect\eqref{eq:model}} (Cosmos-Policy; success \%, LIBERO-Spatial\,/\,LIBERO-Goal). x3: azimuth $\pm45$--$60^\circ$ and elevation 27--$35^\circ$; x4: gains and camera both outside the training range. m1: $x/y$ commands rotated by 30$^\circ$; m3: dead zone 0.15; m4: force ramp to 30\,N.}
\label{tab:supp:extra}
\end{table}

\subsection{Effect of the adapter per coupling}
\label{sec:supp:plugin}
\begin{figure*}[t]
\centering
\includegraphics[width=\textwidth]{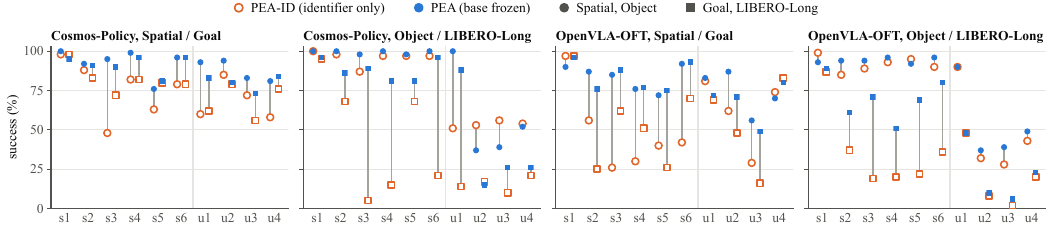}
\caption{\textbf{Effect of the adapter per coupling} ($n{=}100$ per coupling). Open: \pid; filled: \oursfull (base frozen). Means: \cref{tab:main,tab:oft}.}
\label{fig:plugin}
\end{figure*}

\Cref{fig:plugin} compares \pid and \oursfull on every seen-type and unseen-type coupling of both policies and suite pairs.
The adapter helps most where \pid leaves camera-shifted couplings low, \eg, s3 and u1 for Cosmos-Policy and most couplings of OpenVLA-OFT, which was never trained for camera changes.

\subsection{OpenVLA-OFT}
\label{sec:supp:oft}
\Cref{tab:oft} complements \cref{tab:gen} with nominal success and \pid{} w/o drift.
The identifier, applied unchanged from Cosmos-Policy, keeps nominal success and raises new-dynamics success on all four suites; its drift term matters here as for Cosmos-Policy (LIBERO-Spatial: 70.7\% without it, 89.0\% with it).

\begin{table}[t]
\centering
\small
\setlength{\tabcolsep}{2pt}
\resizebox{\linewidth}{!}{%
\begin{tabular}{l c c c c}
\toprule
OpenVLA-OFT & Nominal & Seen & Unseen & New dynamics \\
\midrule
\multicolumn{5}{l}{\textit{LIBERO-Spatial\,/\,LIBERO-Goal}} \\
\raw & \sg{98}{97} & \sg{35.0}{35.3} & \sg{49.0}{43.5} & \sg{63.5}{64.8} \\
\pid{} w/o drift & \sg{99}{99} & \sg{47.7}{55.0} & \sg{61.5}{54.3} & \sg{70.7}{69.8} \\
\pid & \sg{99}{99} & \sg{48.5}{55.2} & \sg{61.5}{54.0} & \sg{\textbf{89.0}}{\textbf{82.3}} \\
\oursfull & \sg{97}{96} & \sg{\textbf{83.7}}{\textbf{84.2}} & \sg{\textbf{74.0}}{\textbf{68.0}} & \sg{84.2}{79.8} \\
\midrule
\multicolumn{5}{l}{\textit{LIBERO-Object\,/\,LIBERO-Long}} \\
\raw & \sg{97}{97} & \sg{78.7}{22.2} & \sg{40.0}{13.3} & \sg{62.2}{55.5} \\
\pid & \sg{99}{99} & \sg{91.8}{36.8} & \sg{48.3}{19.5} & \sg{\textbf{84.8}}{\textbf{77.3}} \\
\oursfull & \sg{95}{97} & \sg{\textbf{94.2}}{\textbf{70.2}} & \sg{\textbf{53.8}}{\textbf{21.8}} & \sg{82.8}{69.3} \\
\bottomrule
\end{tabular}}
\caption{\textbf{OpenVLA-OFT with nominal success} (success \%, means over configurations). The adapter adapts the 224 linear layers of the frozen Llama backbone. Bold: best per block and column, nominal excluded. Per configuration: \cref{fig:supp:oft}.}
\label{tab:oft}
\end{table}

\subsection{LIBERO-Object and LIBERO-Long}
\label{sec:supp:ol}
On the second suite pair, the same trends hold for Cosmos-Policy (\cref{tab:ol}): the identifier raises new-dynamics success on both suites, and the adapter, trained on the LIBERO-Object and LIBERO-Long data, adds most on LIBERO-Long, where the camera shifts hurt the identifier alone most.

\begin{table}[t]
\centering
\small
\setlength{\tabcolsep}{2pt}
\resizebox{\linewidth}{!}{%
\begin{tabular}{l c c c c}
\toprule
Cosmos-Policy & Nominal & Seen & Unseen & New dynamics \\
\midrule
\raw & \sg{100}{98} & \sg{77.7}{33.3} & \sg{52.8}{11.3} & \sg{69.0}{60.3} \\
\pid & \sg{100}{100} & \sg{96.0}{45.3} & \sg{53.5}{15.5} & \sg{\textbf{94.7}}{\textbf{82.5}} \\
\oursfull & \sg{100}{96} & \sg{\textbf{99.3}}{\textbf{88.2}} & \sg{\textbf{57.0}}{\textbf{38.8}} & \sg{91.3}{79.3} \\
\bottomrule
\end{tabular}}
\caption{\textbf{Cosmos-Policy on LIBERO-Object\,/\,LIBERO-Long} (success \%, means over configurations). Unseen types u2--u4 include object masses $\times100$. Bold: best per column, nominal excluded. Per configuration: \cref{fig:supp:ol}.}
\label{tab:ol}
\end{table}

\subsection{Evidence and estimator ablations}
\label{sec:supp:moved}
\Cref{tab:abl} varies, on one model family and dataset, which evidence the policy receives and how it is read.
With an estimator on the FT base, the scripted probe read by the gain-only estimator outperforms the random probe, the 500-step calibration and the ARX identification, and a learned probe encoder adds little on top of it.
Stacking the gain division on a policy that was trained on compensated labels compensates the gain twice and loses 15--20 points on seen types, which is why the adapter trains on intended labels (\cref{sec:method:train}).
Without an estimator, in-episode history, an earlier episode and a random probe all stay near DR-only, and only the scripted probe helps.

\begin{table*}[t]
\centering
\small
\setlength{\tabcolsep}{6pt}
\begin{tabular}{l c c c}
\toprule
Variant (LIBERO-Spatial\,/\,LIBERO-Goal) & Nominal & Seen types & Unseen types \\
\midrule
\multicolumn{4}{l}{\textit{Estimator on the FT base}} \\
scripted probe, gain-only estimator (FT + \gid) & \sg{99}{95} & \sg{\textbf{94.0}}{95.8} & \sg{91.0}{86.5} \\
\quad + learned probe encoder & \sg{98}{97} & \sg{93.8}{\textbf{96.2}} & \sg{\textbf{92.3}}{\textbf{87.8}} \\
random 30-step probe, least squares (FT + ICWM-style) & \sg{97}{94} & \sg{78.5}{81.0} & \sg{72.5}{65.5} \\
500-step calibration, inverse + LMS (FT + SPACE-style) & \sg{79}{65} & \sg{72.0}{68.3} & \sg{65.3}{61.5} \\
scripted probe, ARX + delay + RLS & \sg{90}{69} & \sg{79.0}{59.2} & \sg{73.3}{65.3} \\
500-step calibration, ARX + delay + RLS (FT + ARX) & \sg{95}{83} & \sg{76.3}{61.8} & \sg{85.0}{70.8} \\
\midrule
\multicolumn{4}{l}{\textit{Labels}} \\
compensated labels + gain division (naive stacking) & \sg{99}{97} & \sg{79.0}{75.7} & \sg{75.0}{65.8} \\
\midrule
\multicolumn{4}{l}{\textit{Learned reader without an estimator (compensated labels), by evidence}} \\
none (DR-only) & \sg{96}{96} & \sg{46.2}{52.3} & \sg{55.8}{51.8} \\
in-episode history, 30 steps (Reflective-VLA-style) & \sg{97}{92} & \sg{49.7}{51.7} & \sg{58.5}{55.3} \\
earlier episode (Self-Adaptive-VLA-style) & \sg{95}{95} & \sg{50.0}{56.8} & \sg{54.8}{54.5} \\
random probe, 30 steps (ICWM-style) & \sg{98}{95} & \sg{46.0}{48.5} & \sg{51.3}{52.0} \\
scripted probe (\learn) & \sg{99}{100} & \sg{\textbf{66.3}}{\textbf{68.5}} & \sg{\textbf{61.8}}{\textbf{65.3}} \\
\bottomrule
\end{tabular}
\caption{\textbf{Which evidence, read how?} Controlled comparison on Cosmos-Policy (success \%, means over configurations, $n{=}100$ per configuration; all bases fine-tuned with the same data and schedule). Bold: best per block and column, nominal excluded.}
\label{tab:abl}
\end{table*}

\subsection{LIBERO-RIG}
\label{sec:supp:rig}
On the four overlapping axes of LIBERO-RIG~\cite{cui2026liberorig} (actuator gain, field of view, camera rotation and camera position, levels 3 and 5), \oursfull raises the mean success of Cosmos-Policy from 81.0\% to 89.3\% on LIBERO-Spatial and from 77.0\% to 80.6\% on LIBERO-Goal, and \gid reaches 85.1\% and 78.6\% (\cref{fig:supp:rig}).
The gain axis improves most; our \raw values are within 11 points of those reported by LIBERO-RIG on every configuration.

\section{Real-robot experiments}
\label{sec:supp:real}
\paragraph{Setup.}
We use an AgileX PiPER arm (six joints and a parallel gripper) with a fixed top-down camera and a wrist camera.
The policy is $\pi_{0.5}$~\cite{intelligence2025pi05}, fine-tuned on teleoperated demonstrations of placing one cube beside another.
The actuation shift is a 200\,g payload attached to the end effector; camera and scene are unchanged, so the shift is invisible in the first frame.
We run the full \ours: it probes the arm before each episode, compensates every command and adapts $\pi_{0.5}$ with its LoRA adapter; the base weights stay frozen.
The adapter is trained on episodes that we collected ourselves on this arm, separate from the test trials.

\paragraph{Results.}
Over 10 trials per condition, $\pi_{0.5}$ succeeds in 4 of 10 trials under the payload and \ours in 8 of 10 (95\% Wilson intervals 17--69\% and 49--94\%; Fisher's exact test, two-sided $p{=}0.17$).
\Cref{fig:real} in the main paper shows one rollout per condition.
In these rollouts, the end effector carries the cube at a mean height of 19.4\,cm without payload, 17.0\,cm with the payload and 19.0\,cm with the payload and \ours: the payload lowers the cube so that it strikes the target, and \ours restores the nominal height.

\section{Rendered examples}
\label{sec:supp:qual}
\begin{figure*}[t]
\centering
\includegraphics[width=\textwidth]{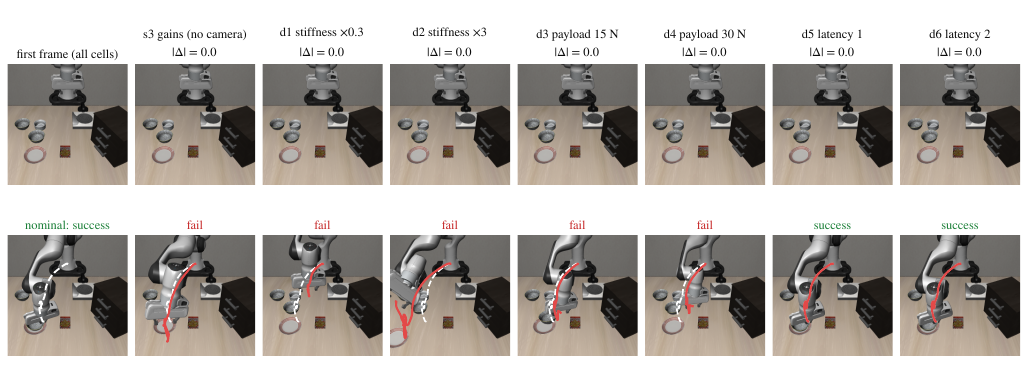}
\caption{\textbf{Actuation shifts are invisible at the start of an episode.} Top: first agent-view frame per configuration; $|\Delta|$ is the mean absolute pixel difference to the nominal frame. Bottom: last frame after replaying the same demonstration open loop; white dashed: nominal end-effector path, red: path under the shift. The replay fails under the s3 gains, both stiffness shifts and both payloads; latency alone (d5, d6) only delays an open-loop replay.}
\label{fig:supp:qual:invisible}
\end{figure*}

\begin{figure*}[t]
\centering
\includegraphics[width=\textwidth]{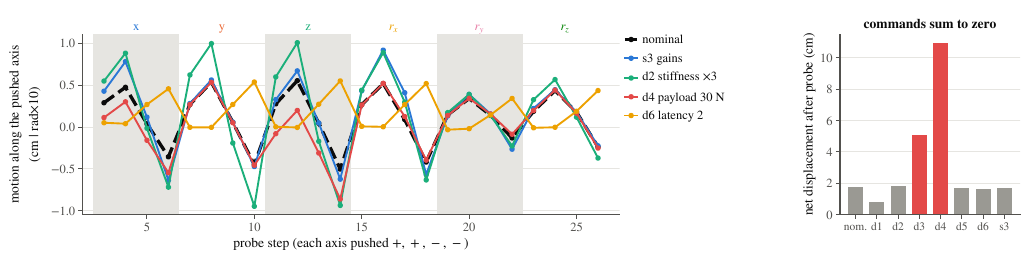}
\caption{\textbf{What the probe reads.} Left: per-step motion along the pushed axis during the probe (cm for translation, rad$\times$10 for rotation). Gains and stiffness scale the response, the 30\,N payload biases it, and a 2-step latency shifts it by two steps. Right: net end-effector displacement at the end of the probe. The probe's commands sum to zero, so the 15\,N and 30\,N payloads (d3, d4) leave 5.1 and 10.9\,cm, against 0.8--1.8\,cm for the other configurations.}
\label{fig:supp:qual:probe}
\end{figure*}

\begin{figure*}[t]
\centering
\includegraphics[width=\textwidth]{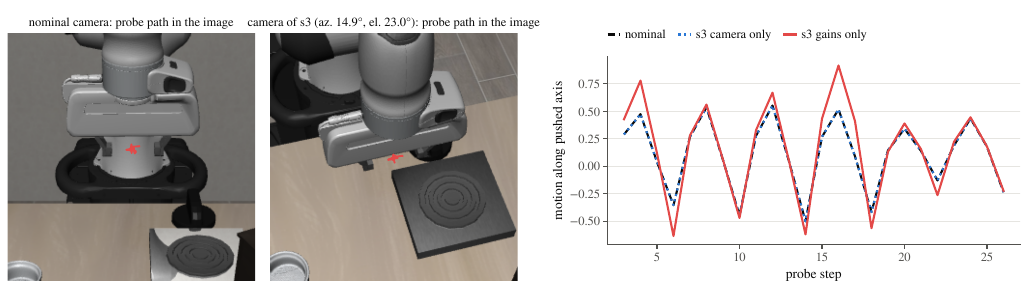}
\caption{\textbf{A moved camera changes how the probe looks, not what it measures.} Left: probe end-effector path (red) projected into the nominal camera and into the camera of s3. Right: motion along the pushed axis in the arm frame. With the s3 camera alone, it coincides with the nominal response; with the s3 gains alone, it does not. The identifier reads the arm-frame response and leaves the camera to the adapter (\cref{sec:method:lora}).}
\label{fig:supp:qual:camera}
\end{figure*}

\Cref{fig:supp:qual:invisible,fig:supp:qual:probe,fig:supp:qual:camera} are rendered in LIBERO with the shifts and the 30-step probe of our evaluation, on LIBERO-Spatial task~0 from the initial state of its first demonstration.
They replay that demonstration open loop or run the probe and are therefore not policy rollouts.

\section{Per-configuration results}
\label{sec:supp:cells}
\begin{figure*}[t]
\centering
\includegraphics[width=\textwidth]{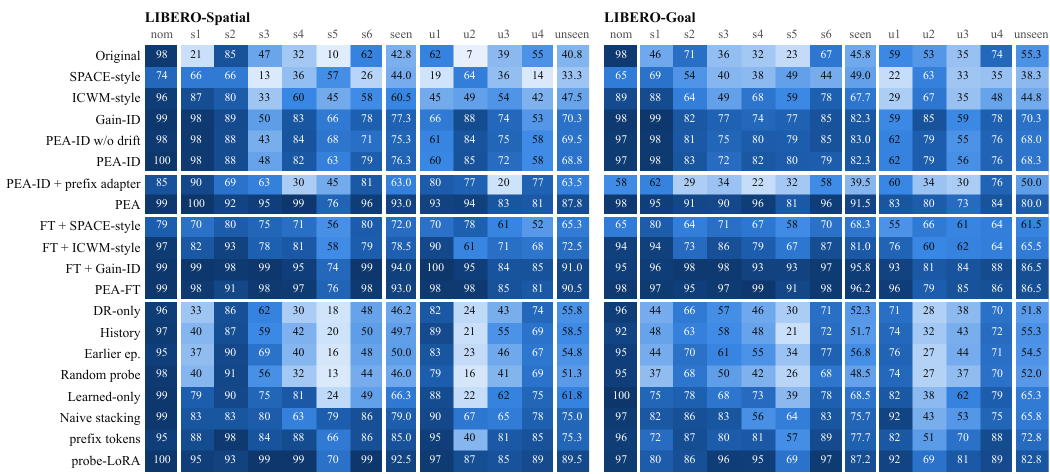}
\caption{\textbf{Per-configuration success on LIBERO-Spatial\,/\,LIBERO-Goal} (Cosmos-Policy, \%, $n{=}100$ per cell; seen, unseen: means over s1--s6 and u1--u4). Groups from top: released weights with a training-free estimator; frozen base with an adapter; FT base with an estimator, and \oursfull-FT; base fine-tuned on compensated labels without an estimator (naive stacking: \learn{} plus division by the \gid estimate). \raw nominal: release protocol (500 episodes).}
\label{fig:supp:main}
\end{figure*}

\begin{figure*}[t]
\centering
\includegraphics[width=\textwidth]{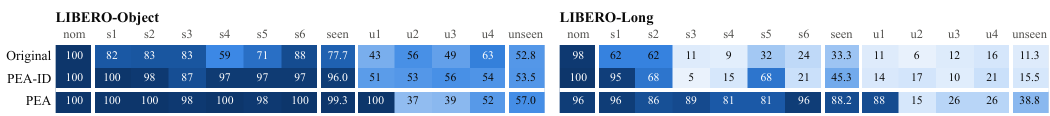}
\caption{\textbf{Per-configuration success of Cosmos-Policy on LIBERO-Object\,/\,LIBERO-Long} (\%, $n{=}100$ per cell; u2--u4 include object masses $\times100$).}
\label{fig:supp:ol}
\end{figure*}

\begin{figure*}[t]
\centering
\includegraphics[width=\textwidth]{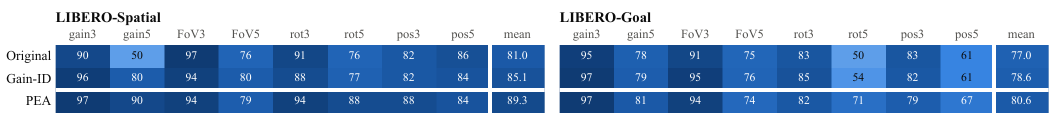}
\caption{\textbf{LIBERO-RIG overlapping axes} (Cosmos-Policy, \%, $n{=}100$ per cell, sign drawn per episode; levels 3 and 5).}
\label{fig:supp:rig}
\end{figure*}

\begin{figure*}[t]
\centering
\includegraphics[width=\textwidth]{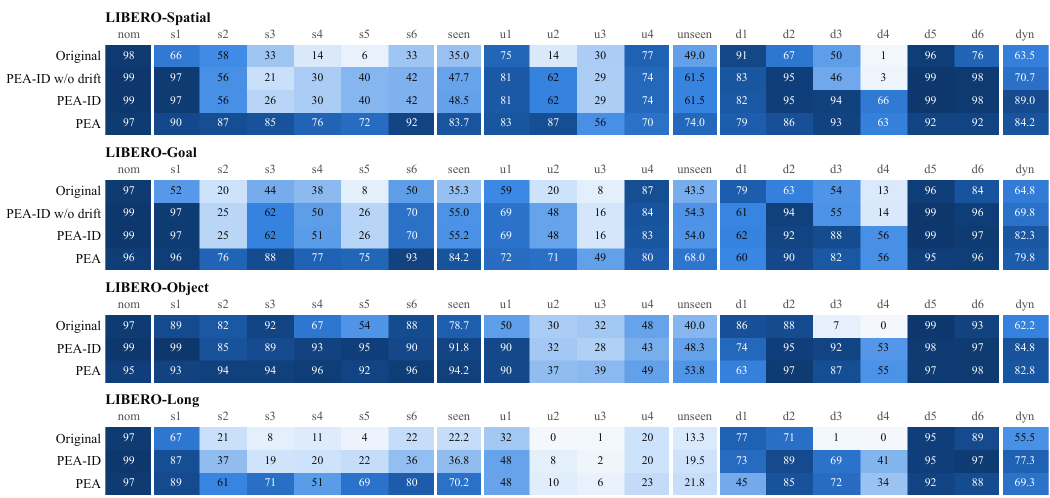}
\caption{\textbf{Per-configuration success of OpenVLA-OFT} (\%, $n{=}100$ per cell). \raw, \pid{} w/o drift (LIBERO-Spatial and LIBERO-Goal) and \pid need no training; \oursfull adds the adapter on frozen weights.}
\label{fig:supp:oft}
\end{figure*}

\Cref{fig:supp:main} gives every configuration behind \cref{tab:main} and \cref{tab:abl,tab:reader}, \cref{fig:supp:dyn} the new dynamics, \cref{fig:supp:ol} LIBERO-Object and LIBERO-Long for Cosmos-Policy, \cref{fig:supp:rig} LIBERO-RIG and \cref{fig:supp:oft} OpenVLA-OFT.

\FloatBarrier

\end{document}